\documentclass[sigconf, nonacm]{acmart}
\usepackage{multirow}
\usepackage{subcaption}
\usepackage{framed}
\usepackage{url}
\usepackage{booktabs} 
\usepackage{amsmath}
\DeclareMathOperator*{\argmax}{argmax}
\usepackage[ruled,vlined,linesnumbered]{algorithm2e}
\SetKw{Continue}{continue}
\SetKw{Break}{break}
\SetKw{Return}{return}
\newtheorem{defn}{Definition}

\usepackage{xcolor}

\copyrightyear{2020} 
\acmYear{2020} 
\setcopyright{acmcopyright}\acmConference[ASE '20]{35th IEEE/ACM International Conference on Automated Software Engineering}{September 21--25, 2020}{Virtual Event, Australia}
\acmBooktitle{35th IEEE/ACM International Conference on Automated Software Engineering (ASE '20), September 21--25, 2020, Virtual Event, Australia}
\acmPrice{15.00}
\acmDOI{10.1145/3324884.3416592}
\acmISBN{978-1-4503-6768-4/20/09}

\begin{document}

\title{Towards Interpreting Recurrent Neural Networks through Probabilistic Abstraction}

 \author{Guoliang Dong}
\affiliation{\institution{Zhejiang University}
  }
  \email{dgl-prc@zju.edu.cn}

  \author{Jingyi Wang}
  \authornote{Corresponding authors: Jingyi Wang and Xinyu Wang.}
  \affiliation{\institution{Zhejiang University}
  }
  \email{wangjyee@zju.edu.cn}

  \author{Jun Sun}
\affiliation{\institution{Singapore Management University}
  }
  \email{junsun@smu.edu.sg}
  
  \author{Yang Zhang}
\affiliation{\institution{Zhejiang University}
    }
  \email{leor@zju.edu.cn}

  \author{Xinyu Wang}
  \affiliation{\institution{Zhejiang University}
  }
  \email{wangxinyu@zju.edu.cn}

  \author{Ting Dai}
\affiliation{\institution{Huawei International Pte Ltd}
    }
  \email{daiting2@huawei.com}

  \author{Jin Song Dong}
\affiliation{\institution{National University of Singapore}
  }
  \email{dongjs@comp.nus.edu.sg}
  
    \author{  Xingen Wang}
  \affiliation{\institution{Zhejiang University}
  }
  \email{newroot@zju.edu.cn}

\begin{abstract}
Neural networks are becoming a popular tool for solving many real-world problems such as object recognition and machine translation, thanks to its exceptional performance as an end-to-end solution. However, neural networks are complex black-box models, which hinders humans from interpreting and consequently trusting them in making critical decisions. Towards interpreting neural networks, several approaches have been proposed to extract simple deterministic models from neural networks. The results are not encouraging (e.g., low accuracy and limited scalability), fundamentally due to the limited expressiveness of such simple models.

In this work, we propose an approach to extract probabilistic automata for interpreting an important class of neural networks, i.e., recurrent neural networks. Our work distinguishes itself from existing approaches in two important ways. One is that probability is used to compensate for the loss of expressiveness. This is inspired by the observation that human reasoning is often `probabilistic'. The other is that we adaptively identify the right level of abstraction so that a simple model is extracted in a request-specific way.
We conduct experiments on several real-world datasets using state-of-the-art architectures including GRU and LSTM. The result shows that our approach significantly improves existing approaches in terms of accuracy or scalability. Lastly, we demonstrate the usefulness of the extracted models through detecting adversarial texts.
\end{abstract}



\maketitle

\section{Introduction}
Neural network models are getting popular due to their exceptional performance in solving many real-world problems, such as self-driving cars~\cite{driving2016cars}, malware detection~\cite{malware2014dtc}, sentiment analysis~\cite{tang2015document} and machine translation~\cite{machine2014translation}. At the same time, neural networks are shown to be vulnerable to issues such as adversarial attacks~\cite{szegedy2013intriguing,goodfellow2014explaining,cw2017Robustness} and embedded back-doors~\cite{backdoor1}. To be able to trust neural networks, it is crucial to understand how neural networks make decisions, even better, to reason about them before deploying them in safety-critical applications.

Neural networks are, however, complex models that work in a black-box manner. Human interpretation of neural networks is often deemed infeasible~\cite{rudin2019stop, lakkaraju2020fool}. Furthermore, the complexity also hinders analysis through traditional software analysis techniques such as testing and verification. Recently, there have been noticeable efforts on porting established software testing and verification techniques to neural network models. For instance, multiple testing approaches like differential testing~\cite{deepXplore}, mutation testing~\cite{DeepMutation,ouricse19}, and concolic testing~\cite{concolic2018testing} have been adapted to test neural networks. Furthermore, several verification techniques based on SMT solving~\cite{katz2017reluplex}, abstract interpretation~\cite{abs2018interpretation} and reachability analysis~\cite{reachability2018analysis} have also been explored to formally verify neural networks. However, due to the complexity of neural networks, existing approaches often have high cost and/or only work for very limited classes of neural networks~\cite{huang2018safety}. 

Recently, an alternative approach has been proposed. That is, rather than understanding and reasoning about neural networks directly, researchers aim to extract simpler models from neural networks. Ideally, the simpler models would accurately approximate the neural networks and be simple enough such that they are human-interpretable. Furthermore, such models can be subject to automated system analysis techniques such as model-based testing~\cite{dalal1999model}, model checking~\cite{clarke1994model} and runtime monitoring~\cite{monitor}. Several attempts have been made on one particularly interesting class of neural networks called recurrent neural networks (RNN), due to their stateful nature as well as popularity in various domains. In~\cite{omlin1996constructing}, Omlin \emph{et al.} propose to encode the concrete hidden states of RNN into symbolic representation and then extract simple deterministic models from the symbolic data~\cite{jacobsson2005rule}. Followup approaches have been proposed to extract different models like deterministic finite automata (DFA) from RNN~\cite{weiss2017extracting,zhou2018learning}. A recent empirical study~\cite{wang2018empirical} shows that such approaches are useful for capturing the structural information of the RNN and hence helpful for monitoring its decision process.

Existing approaches, however, have either limited accuracy (in the case of ~\cite{weiss2017extracting,zhou2018learning} where simple deterministic models are extracted) or scalability (in the case of ~\cite{DBLP:conf/nips/WeissGY19} where more expressive models are extracted). For instance, the extracted models for the real-world sentiment analysis tasks in~\cite{zhou2018learning} have about 70\% fidelity even on the training data. This is not surprising since simple models like DFA have limited expressiveness compared to neural networks. For instance, the work in~\cite{zhou2018learning} extracts deterministic transitions between symbolic encoding of concrete hidden states in RNN, whereas \emph{RNN learned from real-world data are intrinsically probabilistic}. If we were to improve accuracy by extracting more states and transitions, not only the models are computationally expensive to extract but also the extracted models become uninterpretable.

Towards extracting accurate interpretable models from RNN, we develop a technique of extracting probabilistic finite-state automata (PFA) from RNN in this work. Our work distinguishes itself from existing approaches in two important ways. One is that we extract probabilistic models to compensate for the limited expressiveness of simple deterministic models (compared to that of neural networks). This is inspired by the observation that human reasoning is often `probabilistic'~\cite{oaksford2001probabilistic}, i.e., humans often develop `simple' understanding of complex systems by cutting corners (i.e., low-probabilistic cases). The other is that we do not attempt to generate a single model that approximates an RNN model as accurately as possible, as it often leads to models with many states and transitions which are hard to extract or interpret. Instead, we generate models that are sufficiently accurate as per user-request. For instance, if the user requires to get a model which achieves 90\% accuracy in approximating the RNN, the extracted model would have fewer states than a model extracted with 99\% accuracy in doing that. This is achieved by adaptively identifying the right level of abstraction through clustering. 

Our approach is based on a novel algorithm which combines state-of-the-art probabilistic learning and abstraction through clustering.
Given an RNN model and a training set, we first encode the concrete numerical hidden states of an RNN into a set of clusters. Then, we convert the samples in the training set into a set of symbolic traces, each of which is the sequence of clusters visited by the sample. Afterwards, we apply a probabilistic learning algorithm~\cite{mao} on the symbolic traces to learn a PFA.
Furthermore, given a specific requirement, we apply clustering in a greedy way and automatically determine the right number of clusters (i.e., the level of abstraction), which consequently determines the number of states in the learned PFA. That is, we optimize to balance the complexity of the learned PFA (i.e., the fewer states the better) and its accuracy in approximating the RNN (i.e., the higher the better).

We applied our approach to several RNN models with state-of-the-art architectures for solving artificial and real-world (i.e., sentiment analysis) tasks. The results show that our approach significantly improves existing approaches in terms of either accuracy or scalability and is capable of extracting models which accurately approximate the RNN models. Compared to~\cite{zhou2018learning}, our approach improves the fidelity of the extracted model from below 60\% to over 90\% on average on the real-world datasets. Compared to~\cite{DBLP:conf/nips/WeissGY19} which is limited to small artificial datasets, our approach handles large real-world datasets effectively. Lastly, we demonstrate the usefulness of the extracted models through an important application, i.e., detecting adversarial texts that generated to attack RNN. To the best of our knowledge, this is the first systematic approach for detecting adversarial texts.

We organize the rest of the paper as follows. We provide preliminaries in Section~\ref{sec:pre}. We present our approach in detail in Section~\ref{sec:app} and experiment results in Section~\ref{sec:exp}. We review related work in Section~\ref{sec:re} and conclude in Section~\ref{sec:con}.
 \section{Preliminaries}
\label{sec:pre}

In this section, we review relevant background on recurrent neural networks (RNN) and probabilistic finite-state automata (PFA).\\

\begin{figure}[t]
\centering
\includegraphics[width=.33\textwidth]{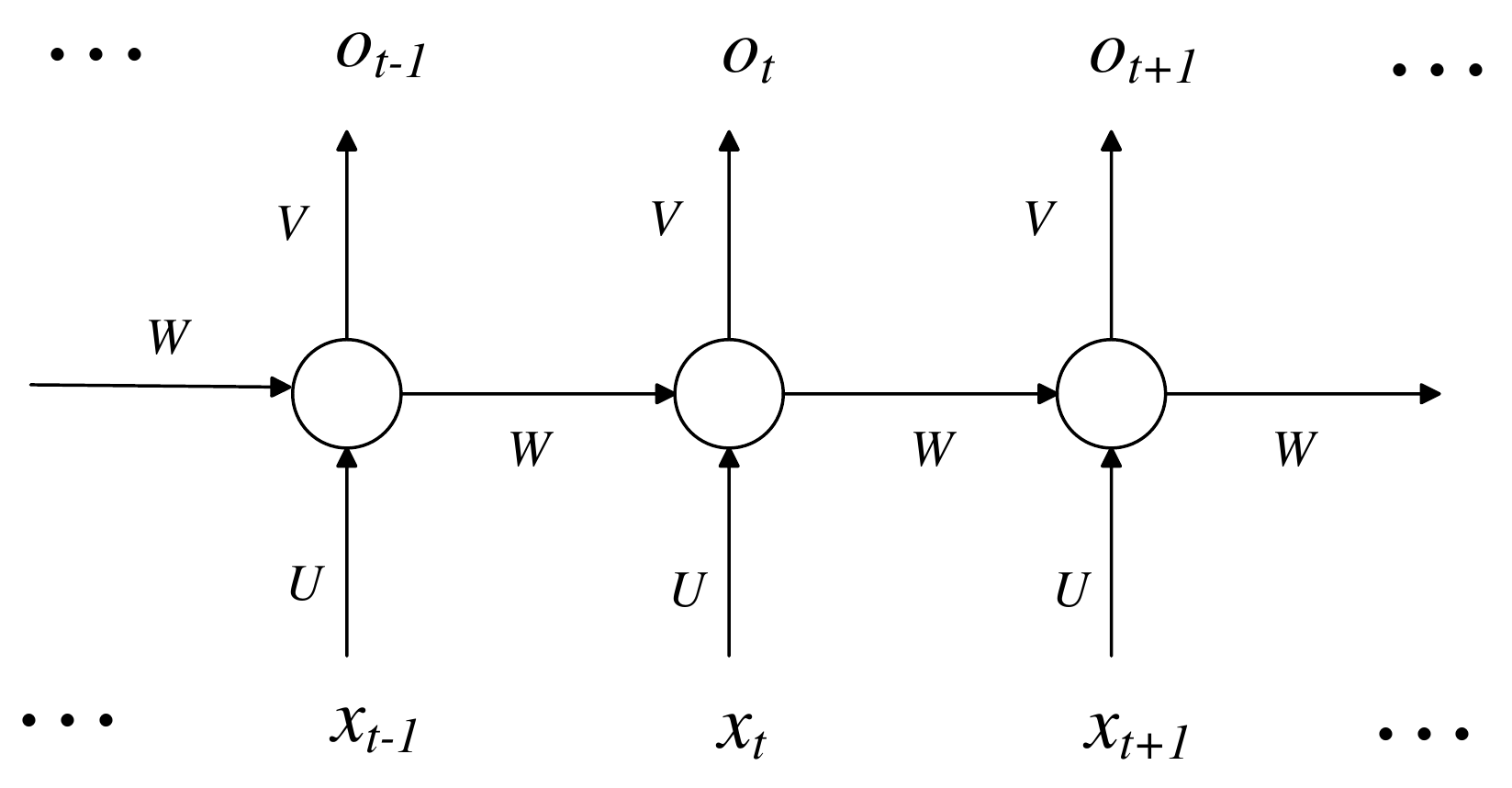}
\caption{A conceptual RNN.}
\label{fig:rnn}
\end{figure}

\noindent \emph{Recurrent Neural Network} In this work, we focus on state-of-the-art RNN architectures such as Gated Recurrent Unit (GRU)~\cite{gru} and Long Short-Term Memory (LSTM)~\cite{lstm}. We introduce RNN at a conceptual level rather than introducing details of GRU and LSTM, since our approach applies to RNN in general. The conceptual model of RNN is shown in Figure~\ref{fig:rnn}, which takes a variable-length sequence $\langle x_0, x_1, \cdots, x_m \rangle$ as input and produces a sequence $\langle o_0, o_1, \cdots, o_m \rangle$ as output. In this work, we focus on RNN classifiers $R:X^*\to I$, where $X$ is the set containing all the possible values of $x$; $X^*$ is the set of finite strings over $X$; and $I$ is a finite set of labels (classification) which only depend on the last output $o_m$.

RNN is stateful, i.e., having a `memory' of previous time steps and what have been calculated so far through a set of hidden states $H$. At each time step $t$, the hidden state $s_t$ and the output $o_t$ are calculated as follows.
\begin{align}
s_{t}=f(Ux_{t} + Ws_{t-1}),\\
o_t=\argmax_{I}\ softmax(Vs_{t}),
\end{align}
where $f$ is usually a nonlinear function like \emph{tanh} or \emph{ReLU}; $U$, $W$, and $V$ are trained parameters; and $softmax$ is a normalizing function which outputs a probability distribution. We remark that GRU and LSTM networks have the same conceptual model shown in Figure~\ref{fig:rnn}, except that more complex functions are used to compute the hidden states. We refer the readers to~\cite{gru,lstm} for details. \\

\noindent \emph{Probabilistic Finite Automata} Existing works on explaining RNN in~\cite{zhou2018learning,weiss2017extracting} focus on extracting models in the form of deterministic finite-state automata (DFA).

\begin{defn}
A DFA is a tuple $\mathcal{A}_D=\langle \mathcal{X}, Q, \delta, Q_0, Q_f \rangle$, where $\mathcal{X}$ is an alphabet; $Q$ is a finite set of all possible states; $\delta:Q\times \mathcal{X}\to Q$ is a labeled transition function; $Q_0 \subseteq Q$ is a set of initial states; and $Q_f \subseteq Q$ is a set of accepting states.
\end{defn}

Note that given a sequence of input symbols, the DFA qualitatively determines whether it is accepted (if the last state is an accepting state) or not. This limits DFA's capability in approximating RNN. For instance, in the context of explaining RNN trained for sentiment analysis, a DFA model produces a binary result (i.e., positive or negative) given any text, whereas RNN often produces a probability of the text being positive or negative. To address this limitation, we instead focus on extracting PFA in this work, which associates probabilities with state transitions in the DFA.

\begin{figure*}[t]
\centering
\includegraphics[width=.65\textwidth]{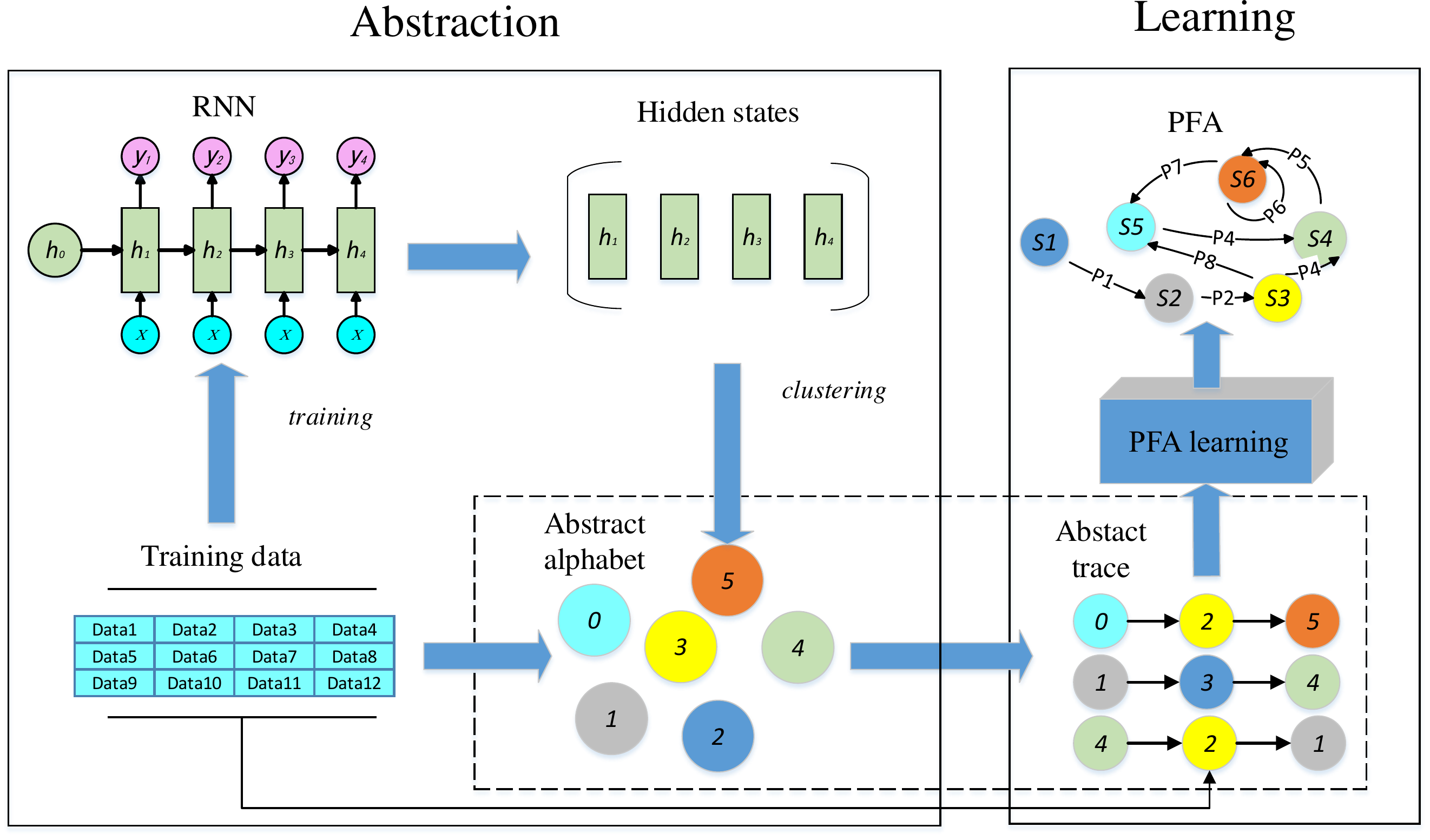}
\caption{Overall framework}
\label{fig:frame}
\end{figure*}

\begin{defn}
A PFA is a tuple $\mathcal{A}=\langle \mathcal{X}, Q, \delta, Q_0, Q_f, \mu_0 \rangle$, where $\mathcal{X}$ is an alphabet; $Q$ is a finite set of states; $\delta:Q \times \mathcal{X}\times Q\to [0,1]$ is a labeled probabilistic transition function such that $\sum_{e \in \mathcal{X}} (\delta(s_i, e, s_j)) = 1$ for all $s_i, s_j \in Q$; $Q_0 \subseteq Q$ is a set of initial states; $\mu_0$ is the initial probability distribution over $Q_0$; and $Q_f$ is a set of accepting states.\end{defn}
An example PFA (extracted using our approach from an RNN model) is shown in Figure~\ref{fig:real-pfa}, where accepting states are represented using double-edged circles; the initial states are indicated with an arrow from nowhere; and each transition is labeled with a probability $p$ followed by a symbol $e$ in the form of $p/e$.

 \section{Our Approach}\label{sec:app}
In this section, we introduce our approach step-by-step.
An overview of the overall workflow is shown in Figure~\ref{fig:frame}. The inputs are an RNN model and the associated training set. There are two main parts, i.e., an abstraction part (on the left) and a learning part (on the right). The abstraction part abstracts the concrete hidden states of a given RNN into an abstract alphabet. The goal is to systematically group hidden states of RNN (in the form of numerical vectors) that exhibit similar behaviors into clusters. The learning part then takes the abstract alphabet and systematically learns a PFA. In the following, we introduce the steps in detail and illustrate them using the following running example.

\begin{example}
\label{exp:1}
The sentence shown in the first row of Table~\ref{tb:exp-1} is a review selected from the RTMR dataset~\cite{rtmr}, which is a widely used movie review dataset for sentiment analysis. The first row in Table~\ref{tb:exp-1} is the original review; the second row is the cleaned input after removing the stop words; and the third row is the label, where 1 represents ``positive'' (i.e., a positive review).

\begin{table*}[t]
\caption{An example input text}
\begin{tabular}{@{}lll@{}}
\toprule
Original review & not a film to rival to live, but a fine little amuse-bouche to keep your appetite whetted &  \\
Cleaned   & film rival live fine little amuse-bouche keep appetite whetted&  \\
Label & 1 \\
Concrete trace & [0.0000, 0.0000, $\cdots$, 0.0000], [-0.0119, -0.0059, $\cdots$, 0.0281], $\cdots$, [-0.0241, 0.1246, $\cdots$, -0.1183]  \\
Abstract trace & $s, 1, 1, 1, 0, 0, 0, 0, 0, 0, P$ \\
\bottomrule
\end{tabular}
\label{tb:exp-1}
\end{table*}
\end{example}
\subsection{Abstraction}
\label{sec:abs}

The objective of this step is to systematically abstract the hidden states of the RNN. Note that the states of the RNN are in the form of numerical vectors which have numerous values and are hard to interpret. The idea is that many of the hidden states represent similar behaviors and thus can be grouped together.
There are two existing techniques on abstracting the hidden states of RNN, i.e., interval partition~\cite{weiss2017extracting} and clustering~\cite{zhou2018learning}. The former arbitrarily partitions the range of hidden state values into multiple intervals and assumes that those hidden states in the same interval have similar behaviors, whereas the latter assumes that nearby hidden states exhibit similar behaviors. In this work, we choose the latter for two reasons. Firstly, clustering has been shown to be intuitive and effective in a recent empirical study~\cite{wang2018empirical}. In fact, it has been shown that the hidden states naturally form clusters~\cite{zhou2018learning}. Secondly, existing clustering techniques allow us to flexibly control the level of abstraction by controlling the number of clusters. 

The goal of clustering is to group the `infinite' hidden state space of RNN into a few clusters. There are many existing clustering algorithms~\cite{xu2005survey}. In this work, we adopt the K-Means algorithm. The idea of K-Means is to identify an assignment function $C:H\to K$ where $H$ is a set of concrete states and $K$ is a set of clusters. Intuitively, $C(h) = k$ means that the concrete state $h$ is mapped to cluster $k$. Ideally, the hidden states that are assigned to the same cluster should be close to each other in terms of certain distance metrics (e.g., Euclidean distance).
Assume that there are $K$ clusters, the assignment $C$ can be found by optimizing the following objective.
\begin{equation}
C^* = \arg\min_C{\sum_{k=1}^{K}N_k\sum_{h\in C_k}||h-\bar{h}_k||^2}
\end{equation}
where $C_k = \{h|h\in H, C(h)=k\}$ is the set of hidden states which are assigned to the cluster $k$; $N_k$ is the size of $C_k$; and $\bar{h}_k$ is mean of all the hidden states in $C_k$. \\

\noindent \emph{Abstracting traces} Clustering is applied as follows in our work. We first collect the RNN hidden state vectors of each sample in the training set. Next, we train a clustering function using the K-means algorithm on those vectors. Note that the number of clusters $K$ is a parameter which must be fixed before applying the K-means algorithm. We discuss how to set the value for $K$ in Section~\ref{sec:modelselection}.

Once we have a clustering function $C$, we obtain the abstract alphabet (i.e., the clusters) and construct a set of abstract traces based on the training set as follows. We feed every sample in the training set into the RNN, and obtain the concrete hidden state at each step. The result is a sequence of concrete hidden states $\langle s_0, s_1, s_2, \cdots, s_n \rangle$ where $s_0$ is a dummy initial state (i.e., the dummy initial mapping). Next, we apply $C$ to each concrete hidden state but the dummy initial state $s_0$ and obtain an abstract trace $\langle s, C(s_1), C(s_2), \cdots, C(s_n) \rangle$ where $s$ is symbol denoting the dummy initial state. Given a sample $x$, we write $\alpha(x)$ to denote the abstract trace obtained as described above.
Afterward, we further concatenate $\alpha(x)$ with the label of $x$ (which forms the accepting states of the extracted PFA as explained later). Note that the abstract alphabet is thus $\mathcal{X}=\{s\}\cup K \cup I$ where $K$ is the set of clusters and $I$ is the set of labels.
We denote the above procedure as $\Psi(x, R, C_K)$ where $R$ is the given RNN, $C_K$ is the clustering function (parameterized by $K$) and $x$ is a sample.
Applying the above procedure to every sample in the training set $X$, we obtain a bag of abstract traces, denoted as $\alpha(X)$, as input for the next phase of our approach.
\begin{example}
\label{exp:2}
Given Example~\ref{exp:1}, let $K$ be 2. The clustering function maps all hidden state vectors to two clusters, denoted as 0 and 1 for simplicity. The abstract alphabet is thus $\mathcal{X}=\{s, 0, 1, P, N\}$ where $P, N$ are labels representing ``positive'' and ``negative'' respectively. The concrete trace obtained from the example text is shown in the fourth row of Table~\ref{tb:exp-1}. With the trained clustering function $C_K$, the abstract trace is shown in the fifth row. \end{example}

\subsection{Learning}
The task of the learning part is to construct a PFA based on the abstract traces.
Ideally, the PFA should be simple (i.e., having a small number of states and transitions) and should have the maximum likelihood of exhibiting the abstract traces (i.e., accurate with respect to the RNN). Our approach is built on top of the \emph{AAlergia} learning algorithm proposed in~\cite{mao}. We choose \emph{AAlergia} as it has been proved to be useful for learning models suitable for system reasoning like probabilistic model checking~\cite{mao}. The key idea of our learning part is to generate a PFA which \emph{generalizes} the probabilistic distribution of the abstract traces over the alphabet. Note that this is fundamentally different from existing approaches such as~\cite{Du:2019:DMQ:3338906.3338954} which uses user-provided partitioned intervals as system states directly.

The details of the learning algorithm are shown in Algorithm~\ref{alg:aa}. The high-level idea is as follows. We first organize the abstract traces into a tree called frequency prefix tree (FPT), which can be considered as a huge model that exhibits exactly the set of abstract traces. Afterwards, we repeatedly merge the nodes in the FPT such that the number of states is reduced gradually. Note that two nodes are merged only if they exhibit similar behaviors. Once all nodes with similar behaviors are merged, we transform the resultant FPT into a PFA. In the following, we present each step in detail.

\begin{figure}[t]
\centering
\includegraphics[width=.37\textwidth]{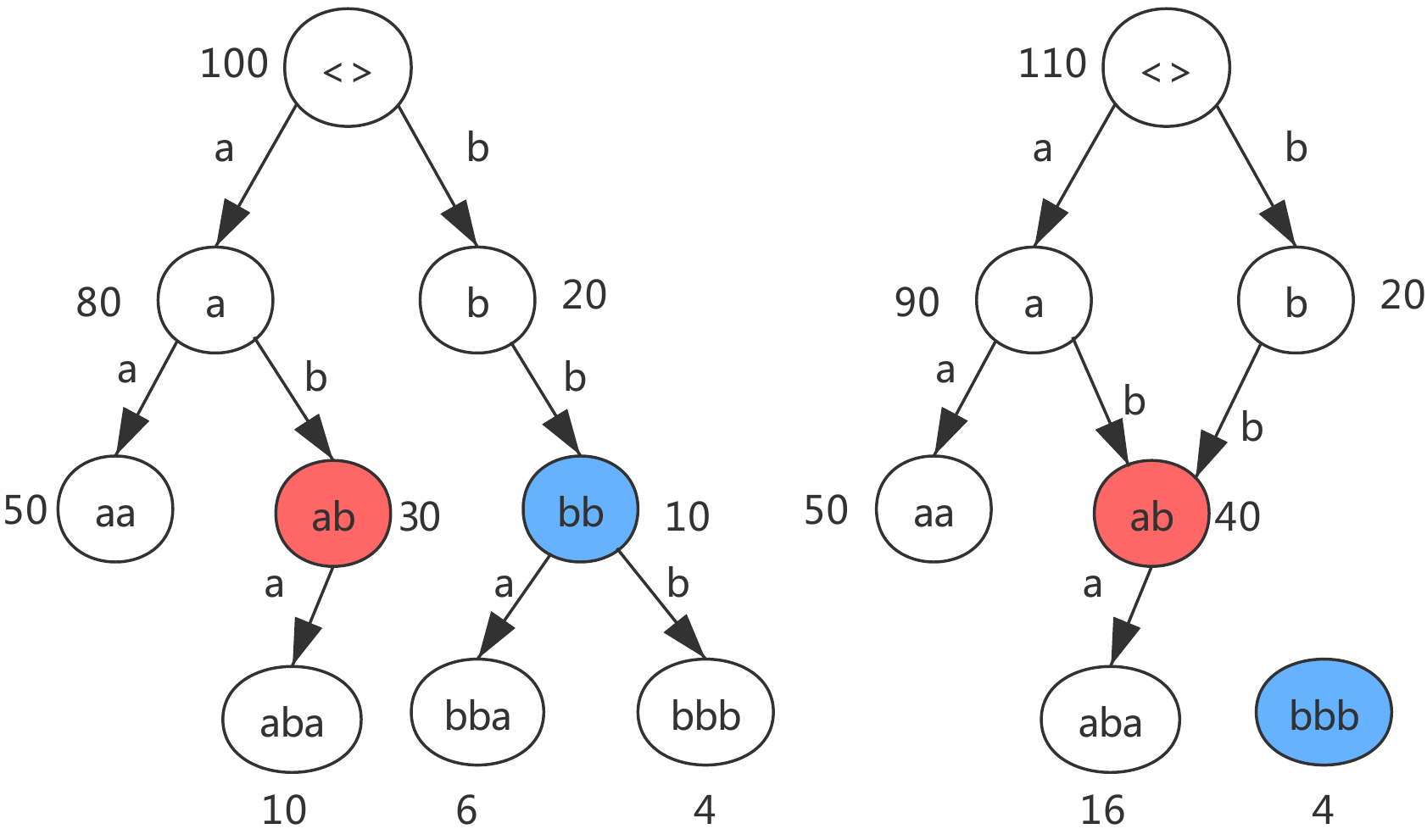}
\caption{Merging nodes}
\label{fig:fp-tree}
\end{figure}

\vspace{1mm}
\noindent \emph{Frequency prefix tree} The first step is to transform the abstract traces into an FPT. Let $\alpha(X)$ be the set of abstract traces and $\mathcal{X}$ be the abstract alphabet. Let $\textit{prefix}(\alpha(X))$ be the set of all prefixes of any $x\in\alpha(X)$. An FPT is a tuple $\textit{tree}(\alpha(X))=\langle N, E, F, \textit{root}\rangle$, where $N$ is $\textit{prefix}(\alpha(X))$; $E\subseteq N\times N$ is the set of edges such that $(n,n')\in E$ if and only if there exists $\sigma\in\mathcal{X}$ such that $n\cdot\sigma=n'$ where $\cdot$ is the concatenation operator; $F$ is a frequency function which records the total number of occurrences of each prefix in $\alpha(X)$; and $\textit{root}$ is the empty string $\langle\rangle$ which corresponds to the dummy initial state $s$. For instance, given a bag of traces with 50 $\langle a,a \rangle$, 20 $\langle a,b \rangle$, 10 $\langle a,b,a \rangle$, 10 $\langle b \rangle$, 6 $\langle b,b,a \rangle$ and 4 $\langle b,b,b \rangle$, the FPT is shown on the left of Figure~\ref{fig:fp-tree}. We remark that a leaf node of the FPT represents a complete trace (which is associated with a label in $I$), whereas an internal node of the FPT is a prefix associated with a certain symbol in $K$ (i.e., a cluster).

Note that an FPT can be regarded as a PFA. That is, the nodes in the FPT can be regarded as states in the PFA and we can obtain the one-step probability from node $n$ to $n\cdot \sigma$ as $P(n, n\cdot \sigma) = F(n\cdot \sigma)/F(n)$ where $F(n)$ is the number of times $n$ appears in $\textit{prefix}(\alpha(X))$. In addition, the probability that a node transits to itself is $P(n, n)=1-\sum_{\sigma \in \mathcal{X}} P(n, n\cdot \sigma)$. However, the FPT is not a good model due to its size. In other words, although the FPT represents the set of abstract states precisely, there is no generalization (a.k.a.~over-fitting). To construct a concise PFA, we generalize the FPT by repeatedly merging the nodes. Intuitively, two nodes are merged if and only if they have similar future behaviors, which is determined through a compatibility test.

\begin{algorithm}[t]
\caption{$\textit{extract}(\alpha(X),\epsilon)$}
\label{alg:aa}
Organize $\alpha(X)$ into a frequency prefix tree $\textit{tree}(\alpha(X))$\;
Let $\mathcal{R}=\emptyset$ be the set of nodes in the final PFA\;
Let $B=\{\textit{root}\}$\;
\While {$B\neq\emptyset$} {
  Select a node $b$ from $B$\;
  Let $merged=false$\;
  \For{each $r\in \mathcal{R}$}
  {
    Test the compatibility between $b$ and $r$\;
    \If{compatible}{
      $merged=true$\;
      Merge $b$ with $r$\;
      \Break\;
    }
  }
  \If{$!merged$}{
    Add $b$ to $\mathcal{R}$\;
  }
  Remove $b$ from $B$ and add the children of $b$\ to $B$\;
}
Let $\delta$ be a probabilistic transition function\;
\For {each $r \in \mathcal{R}$}{
  \For {each $\sigma \in \mathcal{X}$}{
    $\delta(r,\sigma, r\cdot\sigma) = \frac{F(r\cdot\sigma)}{F(r)}$\;
  } 
  $\delta(r,\langle\rangle,r) = 1 - \sum_{\sigma\in\mathcal{X}}P(r,r\cdot\sigma)$\;
} 
Let $Q_0=\{\langle\rangle\}$ which only contains the root node\;
Let $\mu_0$ be the initial distribution which transits to the root node ($\langle\rangle$) with probability 1\;
Let $Q_f$ be the set of leaf nodes in $\mathcal{R}$\;
Construct the PFA as $\langle\mathcal{X},\mathcal{R},\delta,Q_0,Q_f,\mu_0\rangle$.
\end{algorithm}

\vspace{1mm}
\noindent \emph{Compatibility test}
Two nodes are considered compatible (and thus to be merged) if they agree on the last letter, and their future probability distributions are sufficiently similar~\cite{mao}. While the former is easy to check, to check the latter, we compare the differences between the probability of all paths from the two nodes in the FPT and check if the difference is within a certain bound.  Note that the path probability is the product of the one-step probabilities. That is, the probability of a path $\pi=\langle \sigma_1\sigma_2\cdots \sigma_k \rangle$ from a node $n$ is defined as $P(n,\pi)=P(n,n\cdot \sigma_1)\cdot P(n\cdot \sigma_1,\sigma_2)\cdots P(n\cdot \sigma_1\cdots \sigma_{k-1}, \sigma_k)$.

Formally, the future probability distributions of two nodes $n$ and $n'$ are sufficiently similar if and only if for all path $\pi$
\begin{equation}
\label{eq:cmpTest}
\begin{split}
\forall \pi  ,|P(n,n\cdot \pi) - P(n',n' \cdot \pi)|< &\sqrt{6\epsilon log(F(n))/F(n)} + \\
 &\sqrt{6\epsilon log(F(n'))/F(n')}
\end{split}
\end{equation}
where $P(n,n\cdot \pi)$ is the path probability as defined above and $\epsilon$ is a constant coefficient. Note that a larger $\epsilon$ means that more nodes would pass the compatibility test and consequently be merged. In this work, we set $\epsilon$ to be 64 following the empirical results shown in~\cite{mao}. For instance, the node marked $aa$ and the node marked $aba$ shown in Figure~\ref{fig:fp-tree} form a pair of compatible nodes as their last clusters are the same (i.e., $a$) and their future probability distributions are similar (i.e., both with no future paths).

\vspace{1mm}
\noindent \emph{Merging nodes}
In order to systematically identify and merge the nodes in the FPT, we maintain two sets of nodes, i.e., a set of \emph{red} nodes $\mathcal{R}$ (see line 2 in Algorithm~\ref{alg:aa}) which are to be transformed into states in the learned PFA and a set of \emph{blue} nodes $\mathcal{B}$ (see line 3) which are nodes potentially to be merged into those red nodes. Initially, $\mathcal{R}$ is empty and $\mathcal{B}$ only contains the root node.

Next, the loop from line 4 to 15 systematically checks every node in $\mathcal{B}$ to see whether it is compatible with any red node (at line 8). If there is one, the blue node is merged to the compatible red node at line 11. Otherwise, the blue node is turned to a red one and added into $\mathcal{R}$ at line 14. After that, we add the children of the blue node to the blue set at line 15 (unless the blue node is a leaf node).

At line 11, a blue node $b$ is merged into a red node $r$ is as follows. We update the frequency function of both the ancestors and descendant of the red node $r$. In particular, for any of $r$'s ancestor $r_a$, we add its frequency $F(r_a)$ by $F(b)$; and for any of $r$'s descendants $r_d$, let $\pi_d$ be the onwards path from $r$, we add the frequency $F(r_d)$ by $F(b\cdot \pi_d)$. In addition, we add an edge from $b$'s parent to $r$ (since $b$ is now merged into $r$).

For example, Figure~\ref{fig:fp-tree} illustrates how two compatible nodes are merged. On the left is the original FPT, where the node $bb$ and node $ab$ are assumed to be compatible and thus to be merged. On the right is the updated tree after merging, where the frequency of node $ab$ and all its ancestors are updated.

\vspace{1mm}
\noindent \emph{PFA construction} The loop from line 4 to 15 runs until there are no more blue nodes to be merged. Afterwards, we construct the PFA from line 16 to line 24 as follows. The remaining nodes in the FPT (which are all in the red set now) are turned into states in the PFA. The transitions between states in the PFA are then constructed systematically based on the tree edges from line 17 to line 20. Take one red node $r$ as an example. For each $\sigma\in\mathcal{X}$, the transition probability from $r$ to $r\cdot\sigma$ is defined as $P(r,r\cdot\sigma)=F(r\cdot\sigma)/F(r)$ (line 19). The probability of transition to itself is $1-\sum_{\sigma\in\mathcal{X}}P(r,r\cdot\sigma)$ (line 20). The set of initial states only contains the root node, i.e., the empty trace $\langle\rangle$ (line 21). The initial distribution associates probability 1 with the root node (line 22). Note that the set of accepting states in the PFA are exactly the label set $I$ since all the leaf nodes with the same ending letter $s_f \in I$ will be merged as one state as their future distributions are the same (line 23). Finally, we construct the PFA at line 24.

\begin{example}
Central to the conversion of FPT is to build the probabilistic transition function $\delta$ (from line 16 to 20). For the sake of simplicity,  we take the node $a$ and its outgoing transitions in the left FPT in Figure~\ref{fig:fp-tree} as an example to illustrate the conversion. In this FPT,  $\mathcal{X}$ contains two symbols: $[a, b]$,  and
 $\mathcal{R}$ is $[<>, a, b, aa, ab, bb, aba, bba, bbb]$. For node $a \in \mathcal{R}$,  $\delta(a, a, a\cdot a)= F(a\cdot a)/F(a) $ and  $\delta(a, b, a \cdot b)= F(a\cdot b)/F(a)$. According to the left FPT, $F(a)$, $F(a\cdot a)$ and $F(a \cdot b )$  is 80, 50 and 30 respectively. Thus, we can get that the transition probability from $a$ to $aa$ under symbol $a$ is 0.625, the transition probability from $a$ to $ab$ under symbol $b$ is 0.375 and the transition probability from $a$ to itself is 0.
\end{example}

\subsection{Model Selection} \label{sec:modelselection}
Recall that we aim to extract a small model that approximates the RNN accurately (i.e., by making the same decision on as many inputs as possible). The size of the extracted model matters for human interpretation as well as potential tool-based analysis. The number of states in the learned PFA is largely determined by the number of clusters.
Usually, the more clusters we use, the more accurate the extracted model will be (which is evaluated in Section~\ref{sec:effectiveness}).
In the extreme case, if we consider each valuation of the numeric vectors as a cluster, we would have a huge PFA which is perfectly accurate but hardly interpretable. Thus, we do not attempt to generate a model that approximates an RNN model as accurately as possible, as it often leads to models with many states. Instead, we generate models with a user-required level of consistency with the RNN. Such consistency can be measured using fidelity~\cite{zhou2004rule,zhou2018learning}. 

\begin{algorithm}[t]
\caption{$\textit{Overall}(X, R,\gamma_a, \epsilon)$}
\label{alg:overall}
Let $K$ be 2 \;
\While {not time out}{
Obtain the clustering function $C_k$ using K-means\;
$\alpha(X)$ $\leftarrow$ $\Psi(X,R,C_k)$\;
$\mathcal{A}$ $\leftarrow$ $extract(\alpha(X),\epsilon)$\;
Let $\rho=P(\mathcal{A}(x)=R(x)|\mathcal{A})$\;
\If {$\rho \geq \gamma_a$}{
  \Return{ $\mathcal{A}$}\;
}
Increase $K$ by 1\;
}
\Return{$\mathcal{A}$}
\end{algorithm}

Given a user-request in the form of ``generating a model which has a 90\% fidelity compared to the RNN'', the question is then how to determine the `right' number of clusters. Our answer is to gradually increase the number of clusters until a model satisfying the user-request is generated. Our overall algorithm is shown in Algorithm~\ref{alg:overall}, where $X$ is the set of concrete traces (i.e., the sequence of valuations of the hidden feature vectors of the RNN) generated by the samples in the training set; $R$ is the RNN model; $\gamma_a$ is the required fidelity of the extracted model; and $\epsilon$ is the parameter for compatibility testing (refer to Algorithm~\ref{alg:aa}). The output is an extracted PFA $\mathcal{A}$ which satisfies the user-required fidelity.

The algorithm works as follows. We first obtain the clustering function $C_k$ using the K-means algorithm~\cite{arthur2007k} parameterized by $K$ at line 3. Note that $K$ is initially 2 and is increased by 1 each time. Then we apply the procedure $\Psi(X,R,C_k)$ (presented in Section~\ref{sec:abs}) to obtain a bag of abstract traces $\alpha(X)$ at line 4. After that, we extract a PFA $\mathcal{A}$ using Algorithm~\ref{alg:aa} at line 5. We measure the fidelity of the extracted model at line 6. If the fidelity satisfies the requirement, i.e., the condition at line 7 is satisfied, the extracted model is returned at line 8. Otherwise, we increase the number of clusters by 1 at line 9 and start over again.

To obtain a label of a given sample $x$ using the extracted model, we first generate the concrete trace of $x$ given the RNN, i.e., the sequence of valuations of the hidden feature vectors of the RNN. Next, an abstract trace is extracted using the approach presented in Section~\ref{sec:abs}. Note that the abstract trace is in the form of a sequence of letters, each of which represents a cluster except the last one which represents the label. Next, we remove the label from the abstract trace and simulate it on the extracted PFA (i.e., from the initial state of the PFA, for each letter in the abstract trace, we take the corresponding transition of the PFA). Let $s$ be the last state that is reached via the abstract trace. We then apply probabilistic model checking techniques~\cite{clarke1994model} to compute the probability of reaching every label from $s$. We write $P(x,y)$ where $y$ is a label to denote the above-computed probability for label $y$. The label with the largest probability is then generated as the predicated label by the extracted PFA.

\begin{figure}[t]
\centering
\includegraphics[width=.45\textwidth]{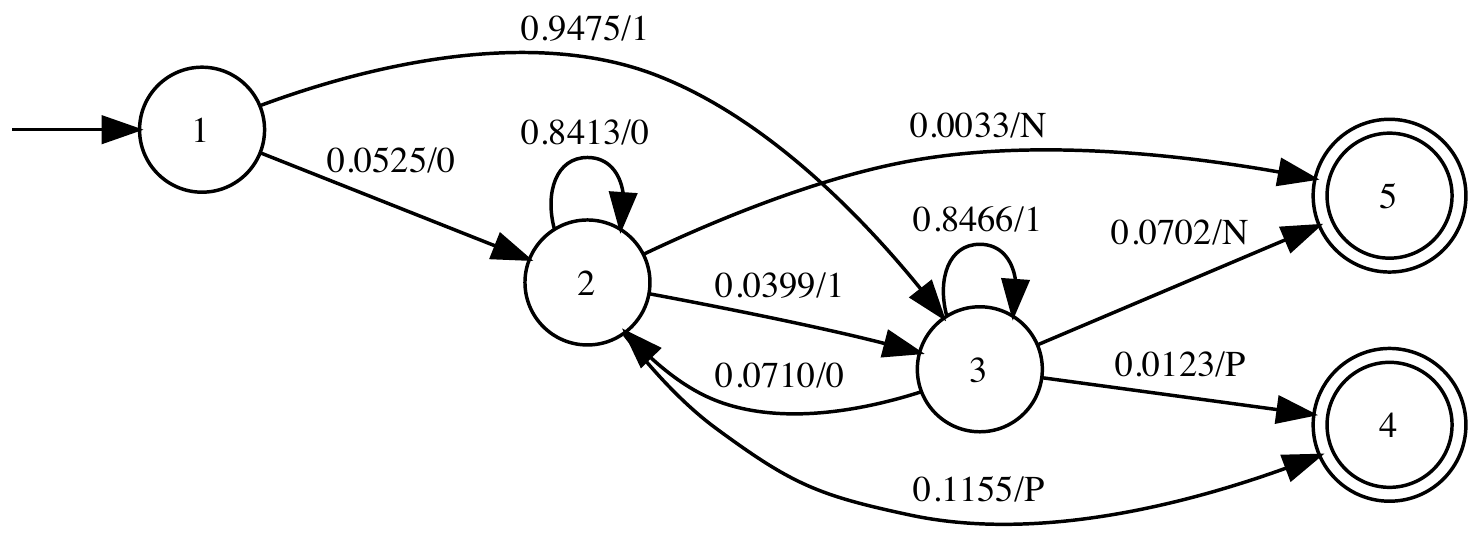}
\caption{Example of a learned PFA}
\label{fig:real-pfa}
\end{figure}

\begin{example}
Figure~\ref{fig:real-pfa} shows the PFA extracted from a GRU model trained on RTMR with 2 clusters. Recall that the abstract trace for the sample text shown in Example~\ref{exp:1} is $\langle s, 1, 1, 1, 0, 0, 0, 0, 0, 0, P\rangle$ (as discussed in Example~\ref{exp:2}). Simulating the trace (excluding the label $P$) on the PFA shown in Figure~\ref{fig:real-pfa}, we end up with state $2$. Applying probabilistic model checking, we obtain that the probability of reaching state 5 (representing label $N$) is 0.1537, whereas the probability of reaching state 4 (representing label $P$) is 0.8463. Thus, the prediction is ``positive''.
\end{example}

We remark that the extracted PFA predicts the label of a sample based on the trace of the RNN, not the sample itself. This is because the PFA is meant to facilitate interpretation of the RNN rather than being a predictive model itself.

 \section{Evaluation}
\label{sec:exp}
Our approach has been implemented as a self-contained prototype, based on Pytorch~\cite{pytorch} and scikit-learn with about 3800 lines of code. Our implementation (including the source code, the dataset, and the trained models) is available at~\cite{guoliang}.

In the following, we evaluate our approach from two aspects. First, we evaluate its effectiveness in terms of extracting PFA, i.e., is it capable of generating small PFA which accurately approximates the RNN? Second, we evaluate its usefulness, i.e., other than being useful for human interpretation, can we use the extracted PFA to solve real-world problems?

\subsection{Effectiveness}
\label{sec:effectiveness}
To evaluate the effectiveness of our approach, we design experiments to evaluate how well the extracted PFA approximate the RNN. Our test subjects are RNN trained on the following datasets.

\begin{itemize}
    \item \emph{Tomita Grammars} is an artificial dataset containing strings generated using different grammars. These grammars were previously adopted for research on RNN model extraction~\cite{weiss2017extracting,wang2018empirical}. They consist of 7 regular languages with different complexity over alphabet $\{0, 1\}$. The detailed definitions of the grammars are listed in Table~\ref{tab:tomita-def}. A string is labeled positive if it is valid according to the grammar. We apply the same setting as in~\cite{weiss2017extracting} to generate a training set and test set based on the grammars. That is, we craft the training set with various lengths for each grammar (i.e., $0$-$13, 16, 19, 22$ except Tomita6 which has a different length setting), and uniformly sample strings of length $1,4,7,\cdots,28$ as the test set for each grammar. The ratio between the training set and the test set is 4 to 1.

    \item \emph{Balanced Parentheses (BP)} is an artificial dataset containing strings generated with an alphabet with 28 letters (i.e., 26 lower-case letters plus `(' and `)'). A string in the dataset is labeled positive if the parentheses in the string are balanced, i.e., each opening parenthesis is eventually followed by a closing parenthesis. Following~\cite{weiss2017extracting}, we generate a set of strings with a length of $0$-$15, 20,25,30$ and a maximum depth of the parentheses 11. Furthermore, the number of positive and negative strings for each length is balanced. The ratio between the training set and the test set is 4 to 1.

    \item \emph{Rotten Tomatoes Movie Review (RTMR)} is a movie review dataset collected from Rotten Tomatoes pages~\cite{rtmr} for sentiment analysis, which contains 5331 positive and 5331 negative sentences. The average length of this dataset is about 21. We take all the samples in the dataset and divide them into two groups, i.e., 80\% as a training set and 20\% as a testing set.

    \item \emph{IMDB} is a widely used benchmark for sentiment analysis classification. It contains $50,000$ movie reviews which are equally split into a training set and a test set. In total, there are 25k positive reviews and 25k negative reviews. The dataset is well collected and processed in a way that makes sure the reviews are as independent as possible. Since the samples in the dataset are much longer than those in RTMR and the size of the dataset is much bigger, to reduce the experiment time, we take those samples with a length less than 51. We also keep 80\% of the selected data as the training set and 20\% as the test set.
\end{itemize}
Table~\ref{tab:reg-dataset} summarizes the size of all datasets. We train RNN models to classify the strings in each dataset. We adopt two popular types of RNNs, i.e., LSTM and GRU. We set the dimensions of hidden states and the number of hidden layers for the two RNNs as 512 and 1 respectively as in~\cite{zhang2015character}. When training the models, we use one-hot encoding to encode each character of the artificial dataset, and use word2vec~\cite{mikolov2013efficient} to transform each word of the real-world dataset into a 300-dimensions numerical vector. Table~\ref{tab:rel-dataset} shows the performance of the trained models, all of which is similar to the state-of-the-art performance. In total, we have 20 models.

\begin{table}[t]
\small
\centering
\caption{Tomita grammars}
\label{tab:tomita-def}
\begin{tabular}{|l|l|}
\hline
Grammar & Definition                                                                        \\ \hline
Tomita1                                                   & 1*                                                                                \\ \hline
Tomita2                                                   & (10)*                                                                             \\ \hline
Tomita3                                                   & \begin{tabular}[c]{@{}l@{}}the complement of\\ ((0|1)* 0)*1(11)*(0(0|1)*1)*0(00)*(1(0|1)* )*\end{tabular}       \\ \hline
Tomita4                                                   & words not containing 000                                                          \\ \hline
Tomita5                                                   & \begin{tabular}[c]{@{}l@{}}the number of ``0'' and the number of ``1'' are \\ both even numbers for each string\end{tabular}     \\ \hline
Tomita6                                                   & \begin{tabular}[c]{@{}l@{}}the difference between the number of ``0''\\  and the number of ``1'' is a multiple of 3\end{tabular} \\ \hline
Tomita7                                                   & 0*1*0*1*                                                                          \\ \hline
\end{tabular}
\end{table}

\begin{table}[t]
\centering
\caption{Size of the datasets}
 \label{tab:reg-dataset}
\begin{tabular}{|c|c|c|c|}
\hline
Task&Dataset& Training set & Test set\\ \hline
\multirow{8}{*}{Artificial Dataset}    &Tomita1 & 624         & 156           \\ \cline{2-4}
		           &Tomita2 & 619        & 160     \\ \cline{2-4}
			&Tomita3 & 2898      & 724        \\ \cline{2-4}
			&Tomita4 & 2911 & 727       \\ \cline{2-4}
			&Tomita5 & 2311     & 577  \\ \cline{2-4}
			&Tomita6 & 3791      &947   \\ \cline{2-4}
			&Tomita7 & 2878      & 719        \\ \cline{2-4}
			&BP        &3978      & 995     \\   \hline
\multirow{2}{*}{Real-world Dataset }&RTMR  &8528 &2134 \\ \cline{2-4}
			&IMDB  &3730 &933  \\ \hline
\end{tabular}
\end{table}

\begin{table}[t]
  \centering
\caption{Performance of target models}
\label{tab:rel-dataset}
    \begin{tabular}{|c|c|c|c|c|}
    \hline
    \multirow{2}[4]{*}{Dataset} & \multicolumn{2}{c|}{LSTM} & \multicolumn{2}{c|}{GRU} \\ \cline{2-5}
    & Training set & Test set & Training set & Test set \\
    \hline
    RTMR  & 79.88\% & 77.46\% & 80.96\% & 77.32\% \\
    \hline
    IMDB  & 88.07\% & 84.35\% & 88.36\% & 84.24\% \\
    \hline
    BP    & 99.92\% & 100\% & 99.90\% & 99.90\% \\
    \hline
    Tomita1 & 99.04\% & 98.72\% & 99.20\% & 98.72\% \\
    \hline
    Tomita2 & 99.67\% & 99.38\% & 99.78\% & 99.38\% \\
    \hline
    Tomita3 & 100\% & 99.86\% & 99.90\% & 99.86\% \\
    \hline
    Tomita4 & 99.90\% & 100\% & 100\% & 100\% \\
    \hline
    Tomita5 & 73.47\% & 74.52\%  & 73.73\% & 75.39\% \\
    \hline
    Tomita6 & 65.34\% & 63.99\% & 64.63\% & 66.42\% \\
    \hline
    Tomita7 & 99.51\% & 100.00\% & 99.34\% & 100.00\% \\
    \hline
    \end{tabular}\vspace{-1.5em}
\end{table}

\begin{figure*}[t]
     \begin{subfigure}{\textwidth}
             \centering
             \includegraphics[width=0.24\textwidth]{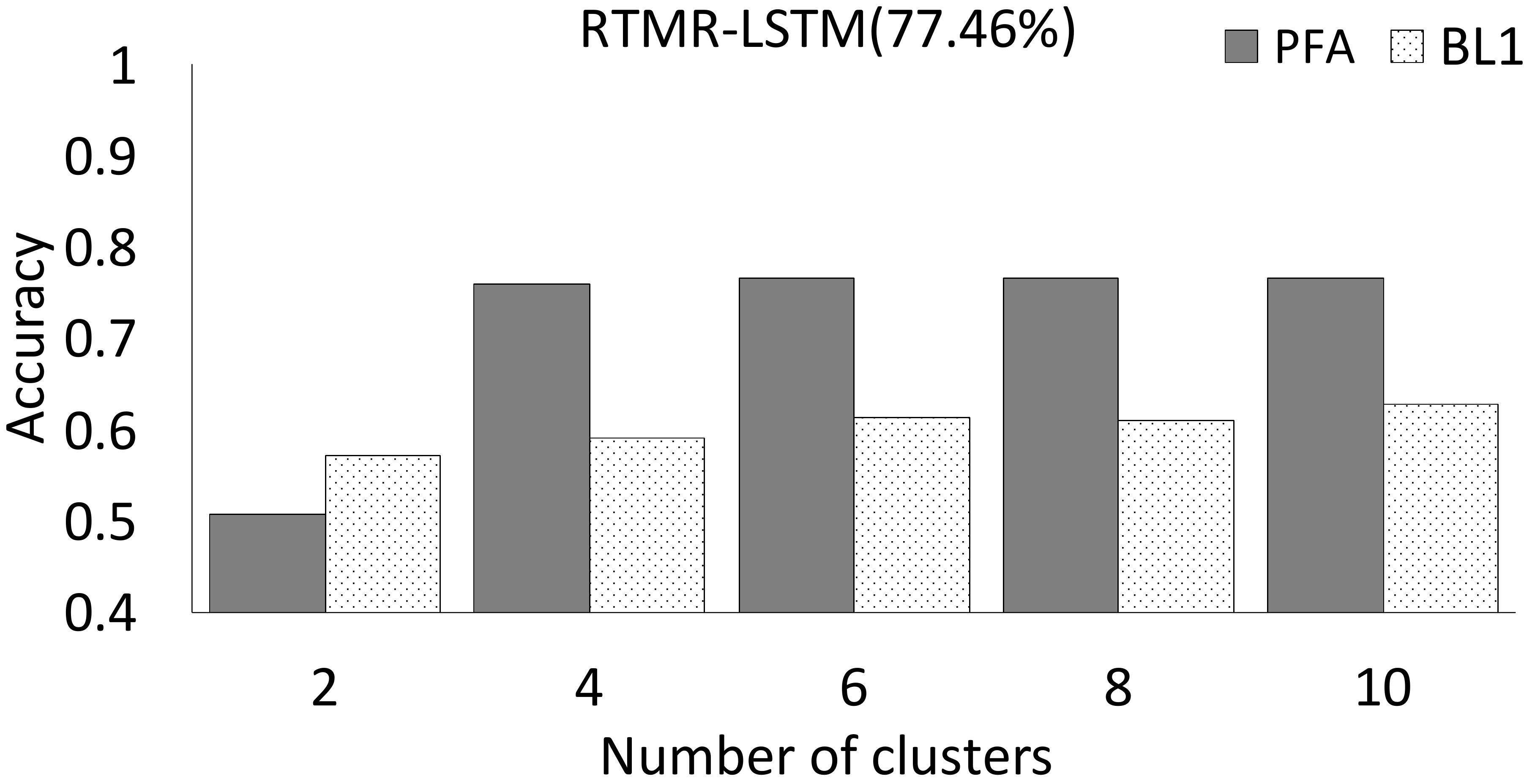} \includegraphics[width=0.24\textwidth]{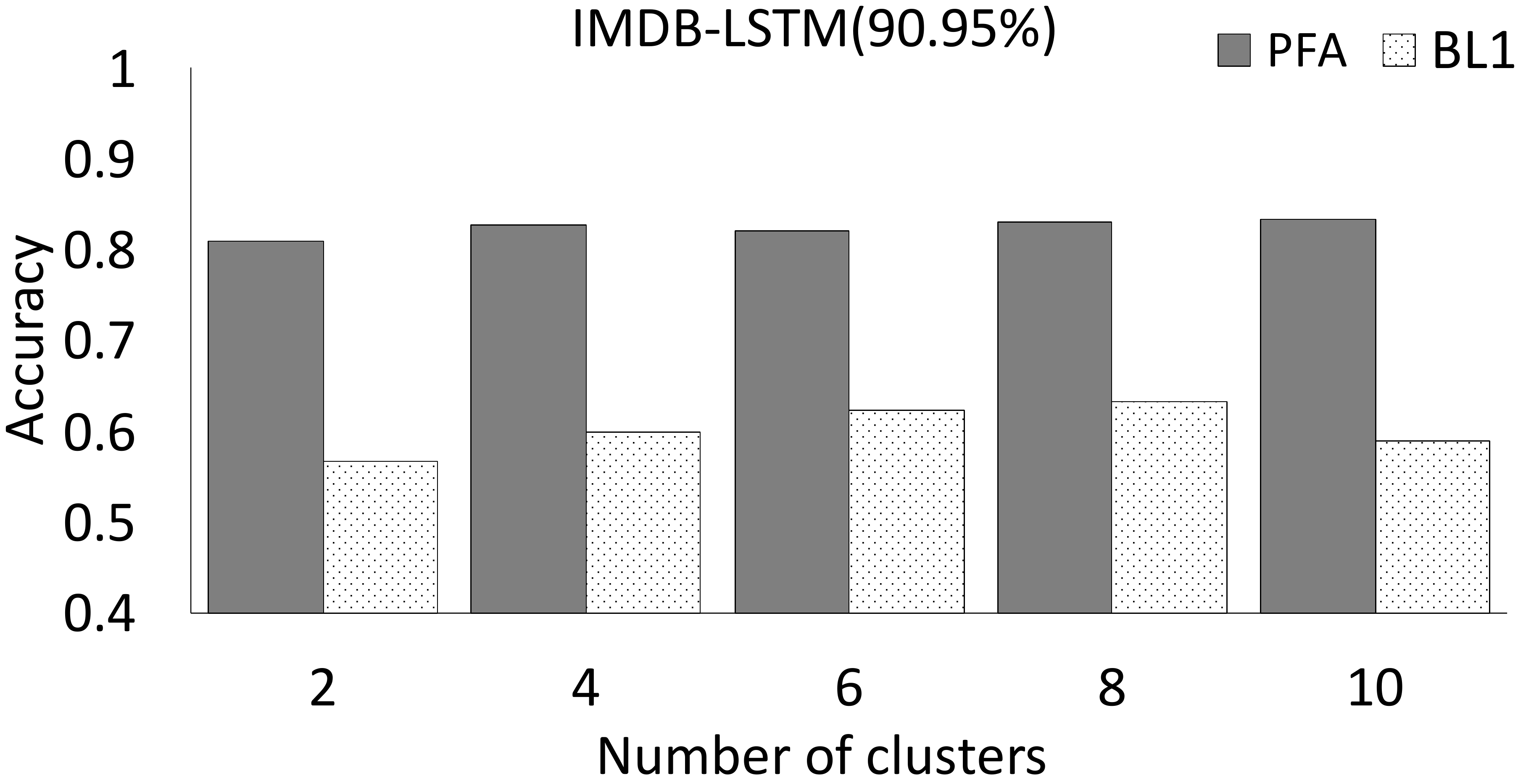} \includegraphics[width=0.24\textwidth]{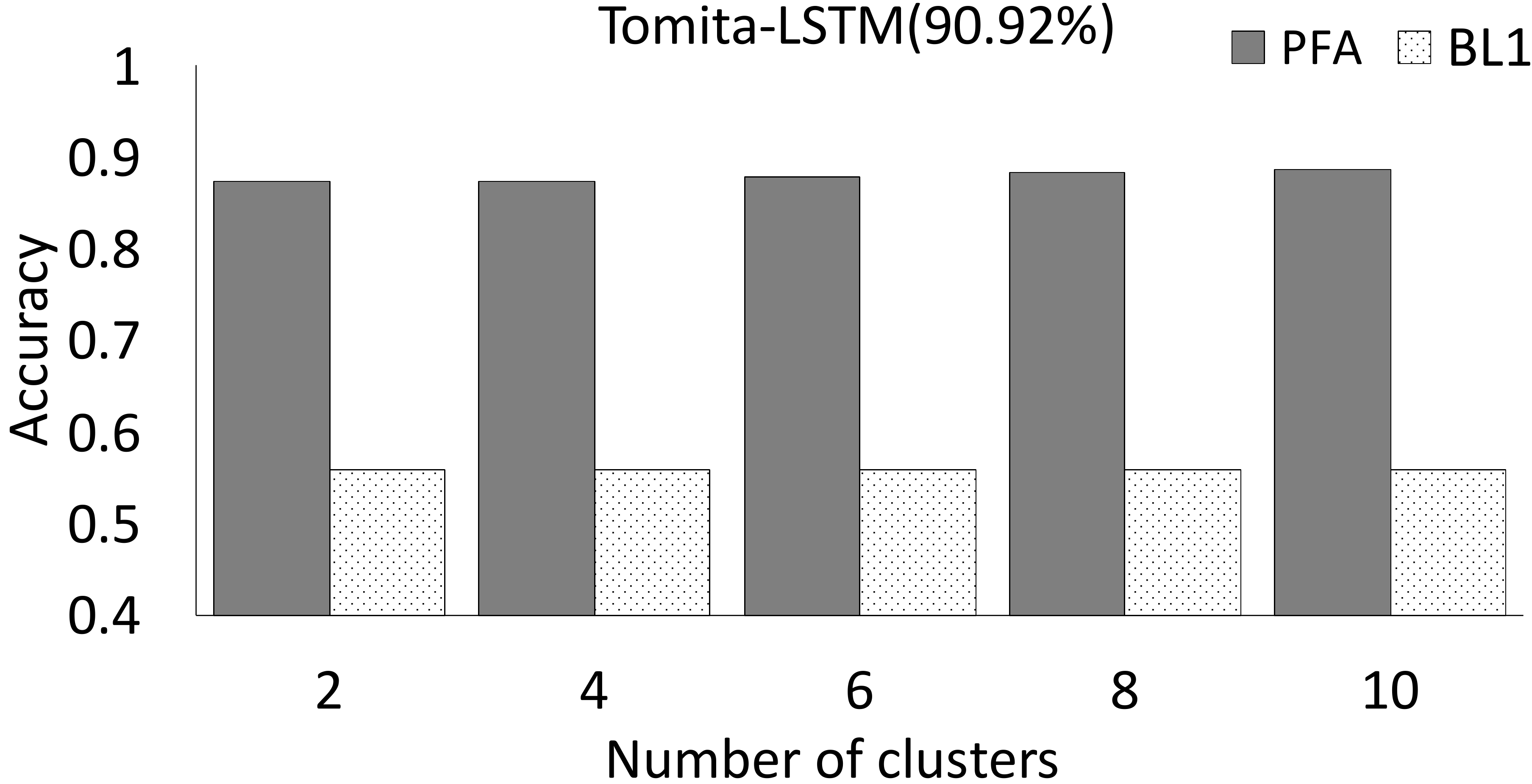} \includegraphics[width=0.24\textwidth]{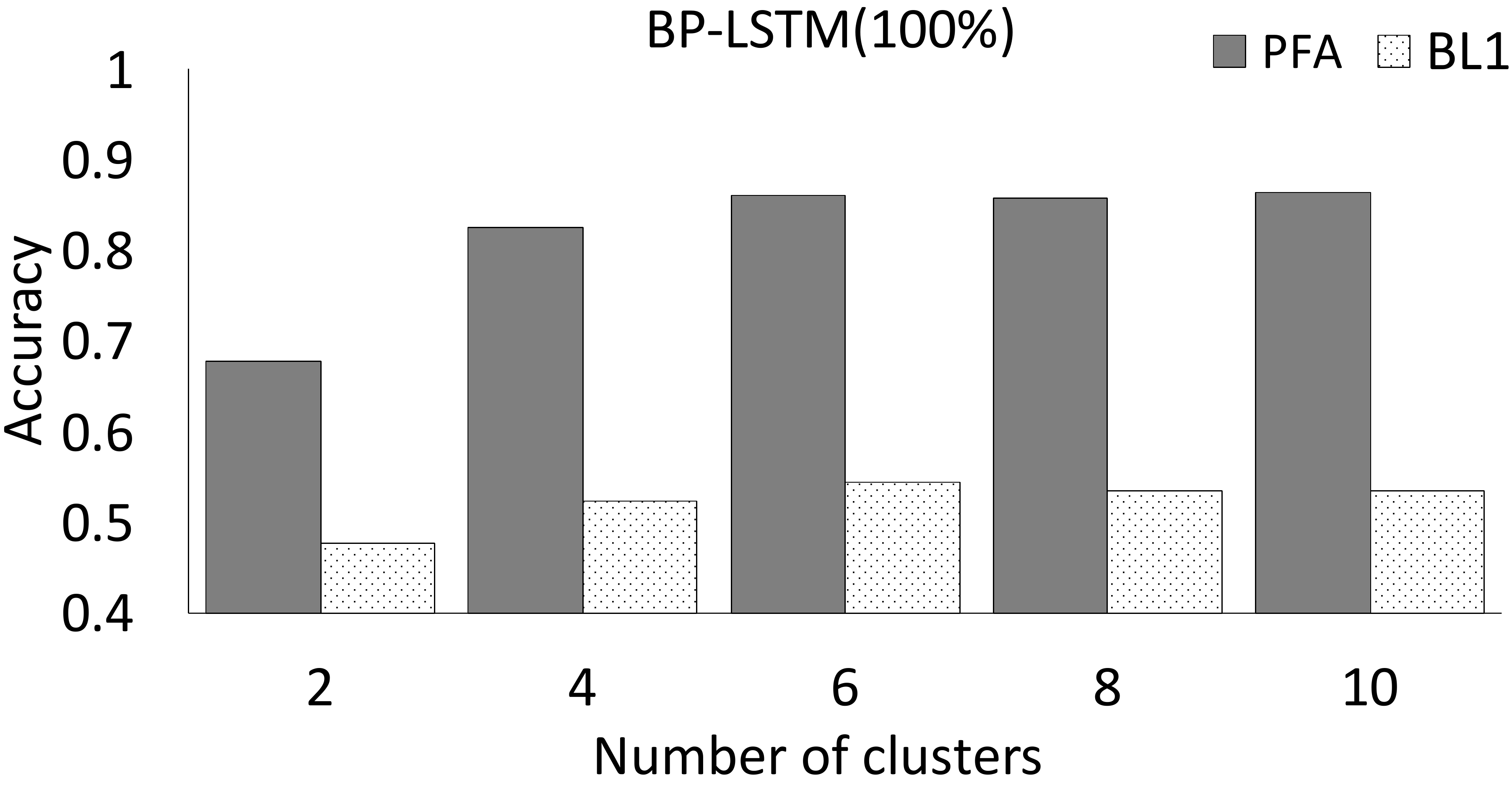} \end{subfigure}
   \begin{subfigure}{\textwidth}
             \centering
             \includegraphics[width=0.24\textwidth]{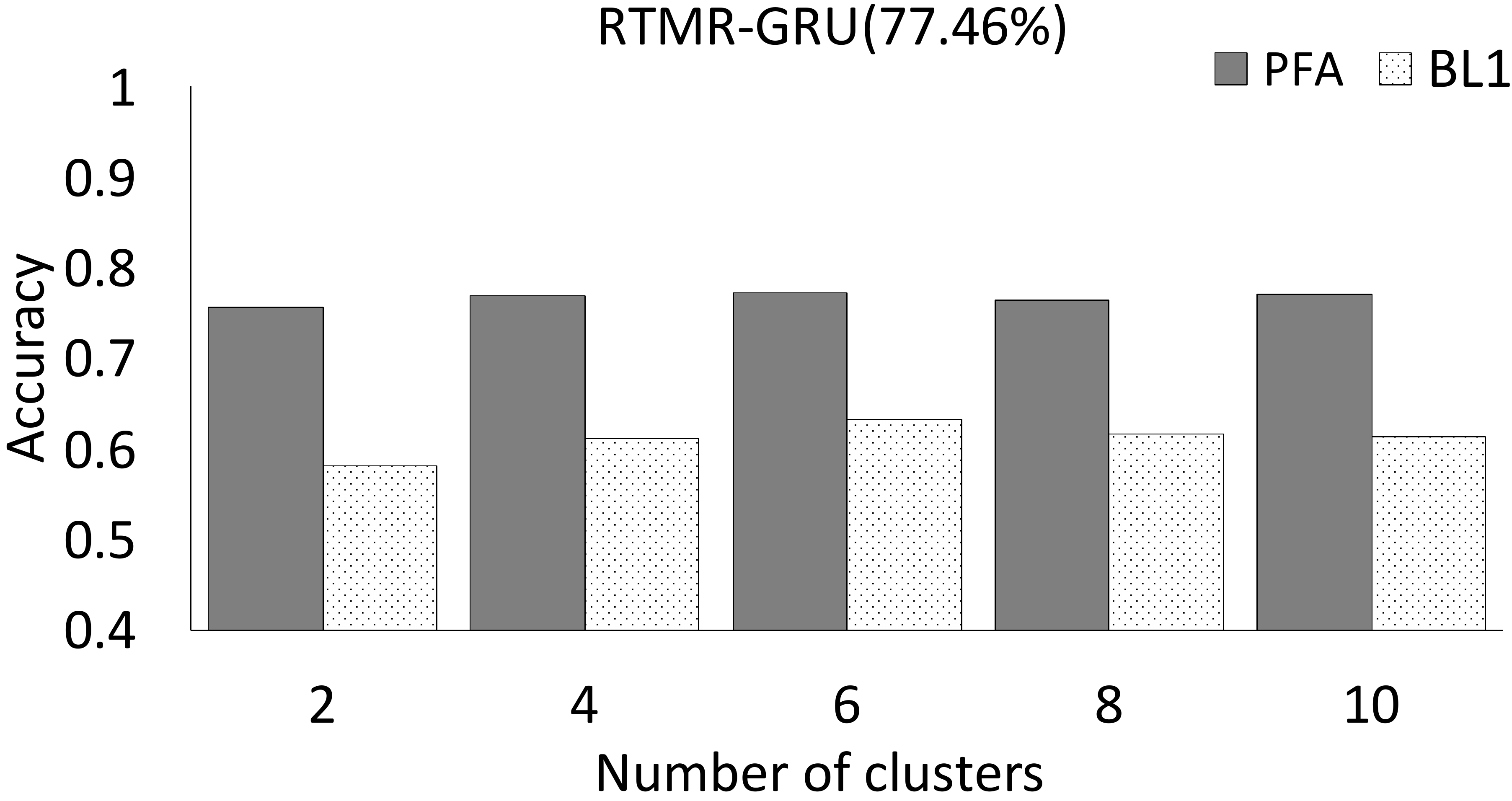} \includegraphics[width=0.24\textwidth]{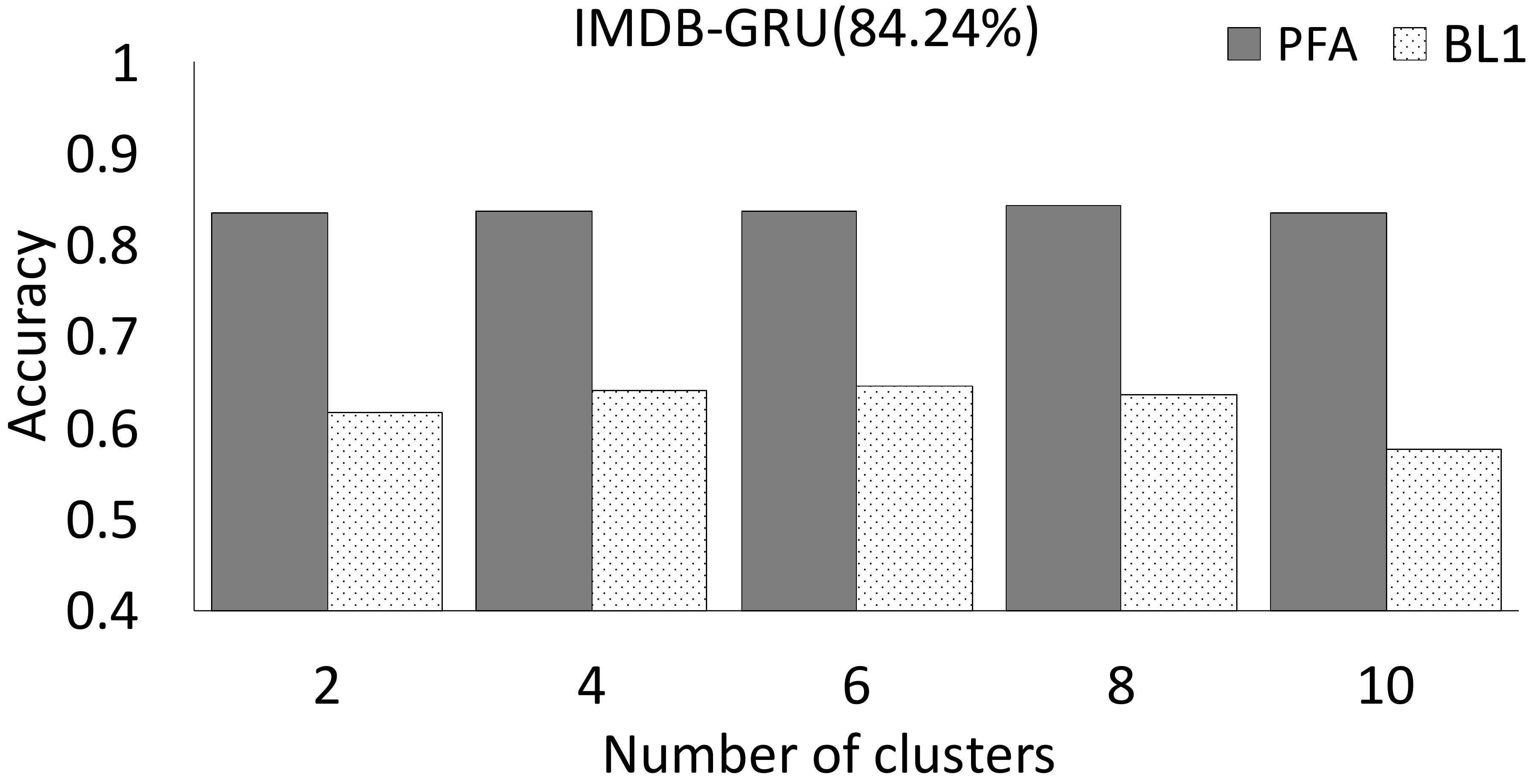} \includegraphics[width=0.24\textwidth]{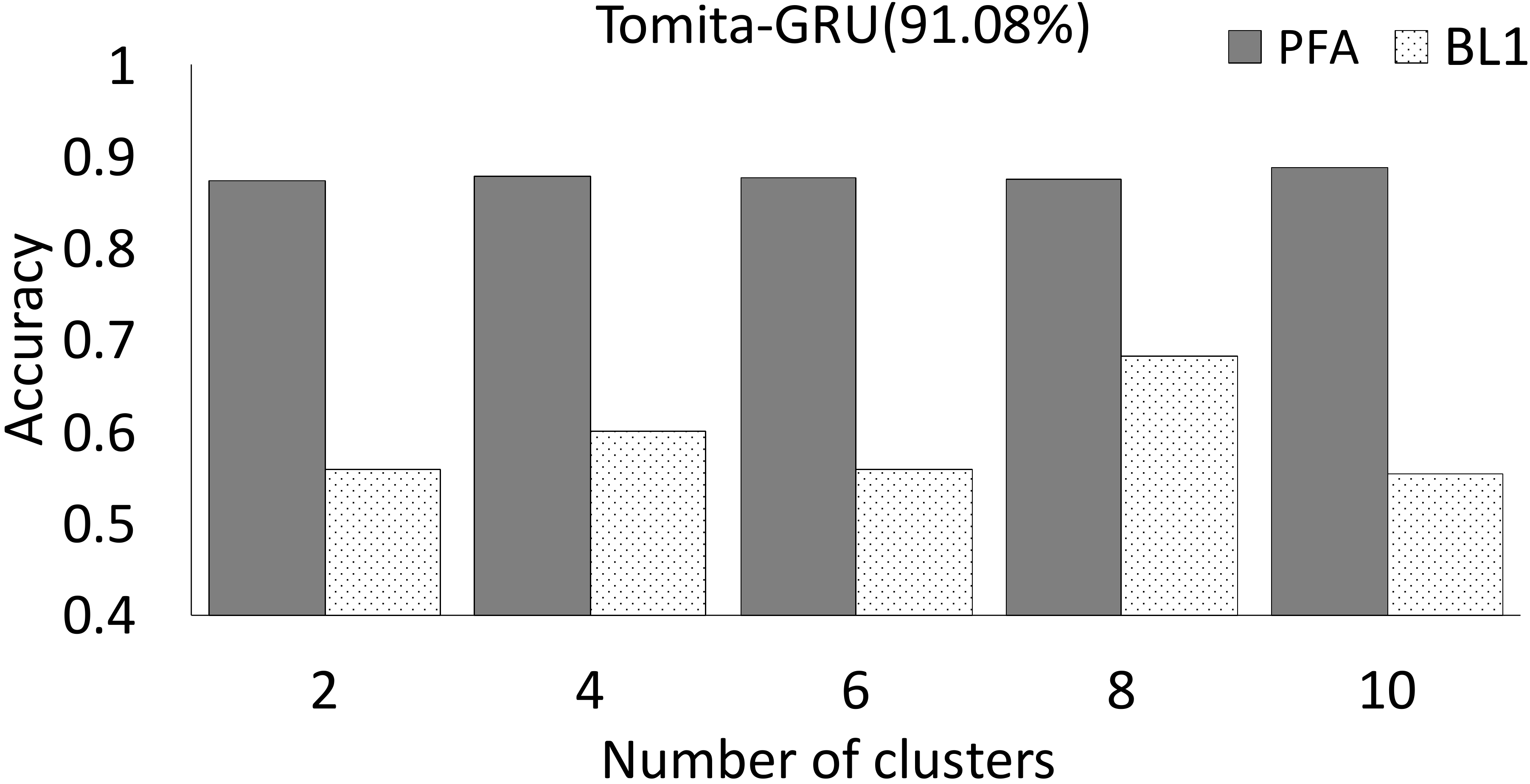} \includegraphics[width=0.24\textwidth]{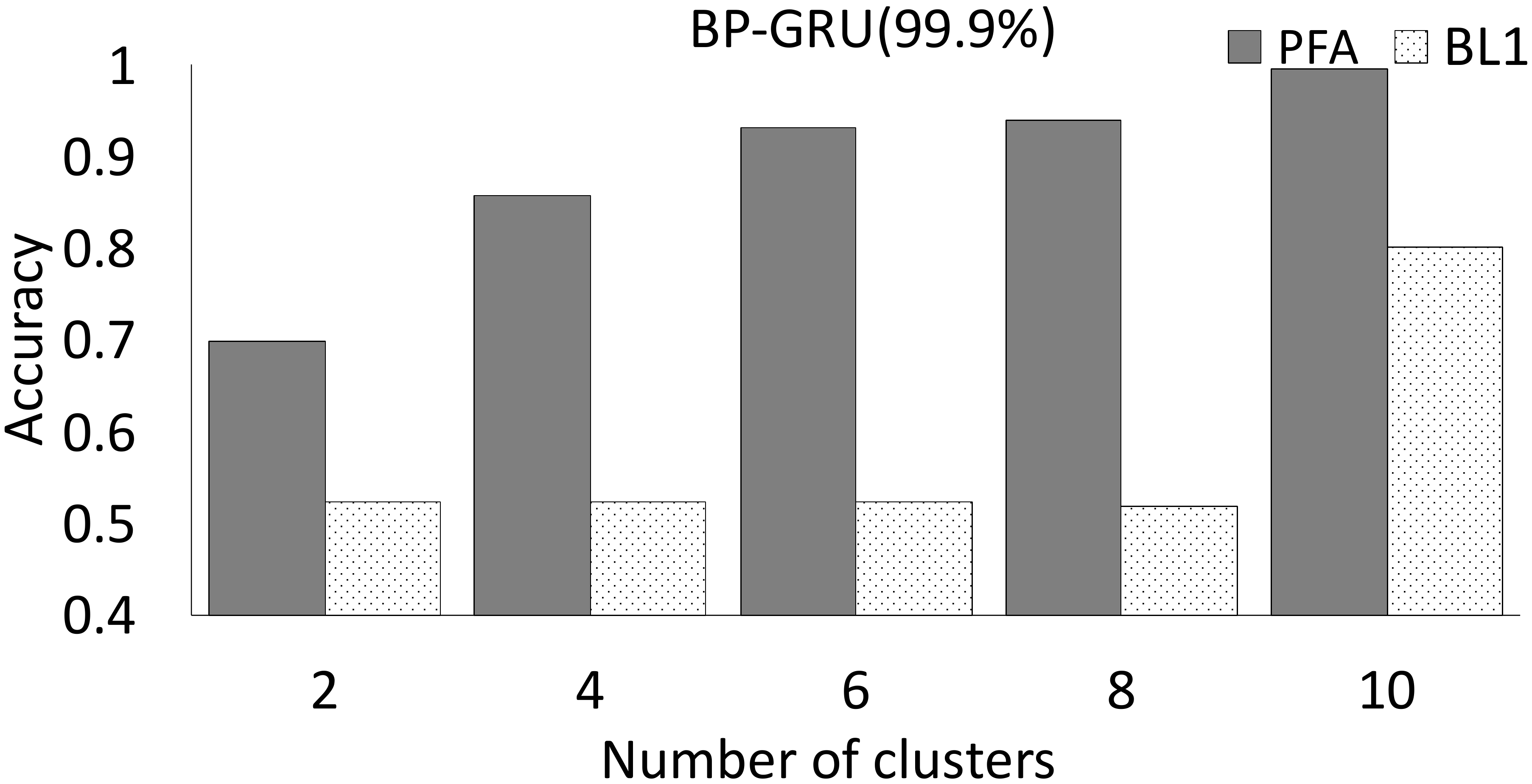} \end{subfigure}
   \caption{The accuracy of our approach vs. BL1}
    \label{fig:acc}
 \end{figure*}

\begin{figure*}[t]
     \begin{subfigure}{\textwidth}
             \centering
             \includegraphics[width=0.24\textwidth]{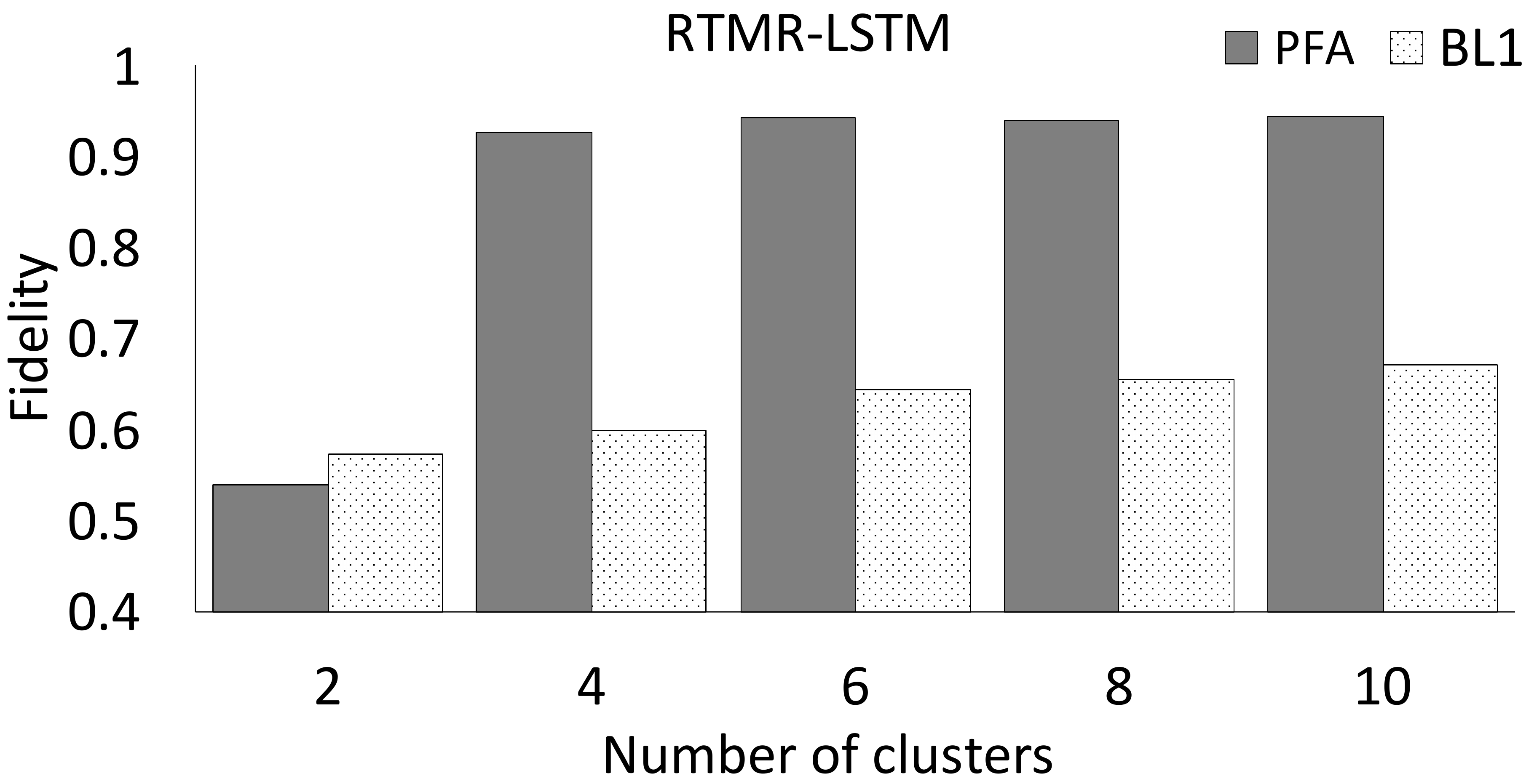} \includegraphics[width=0.24\textwidth]{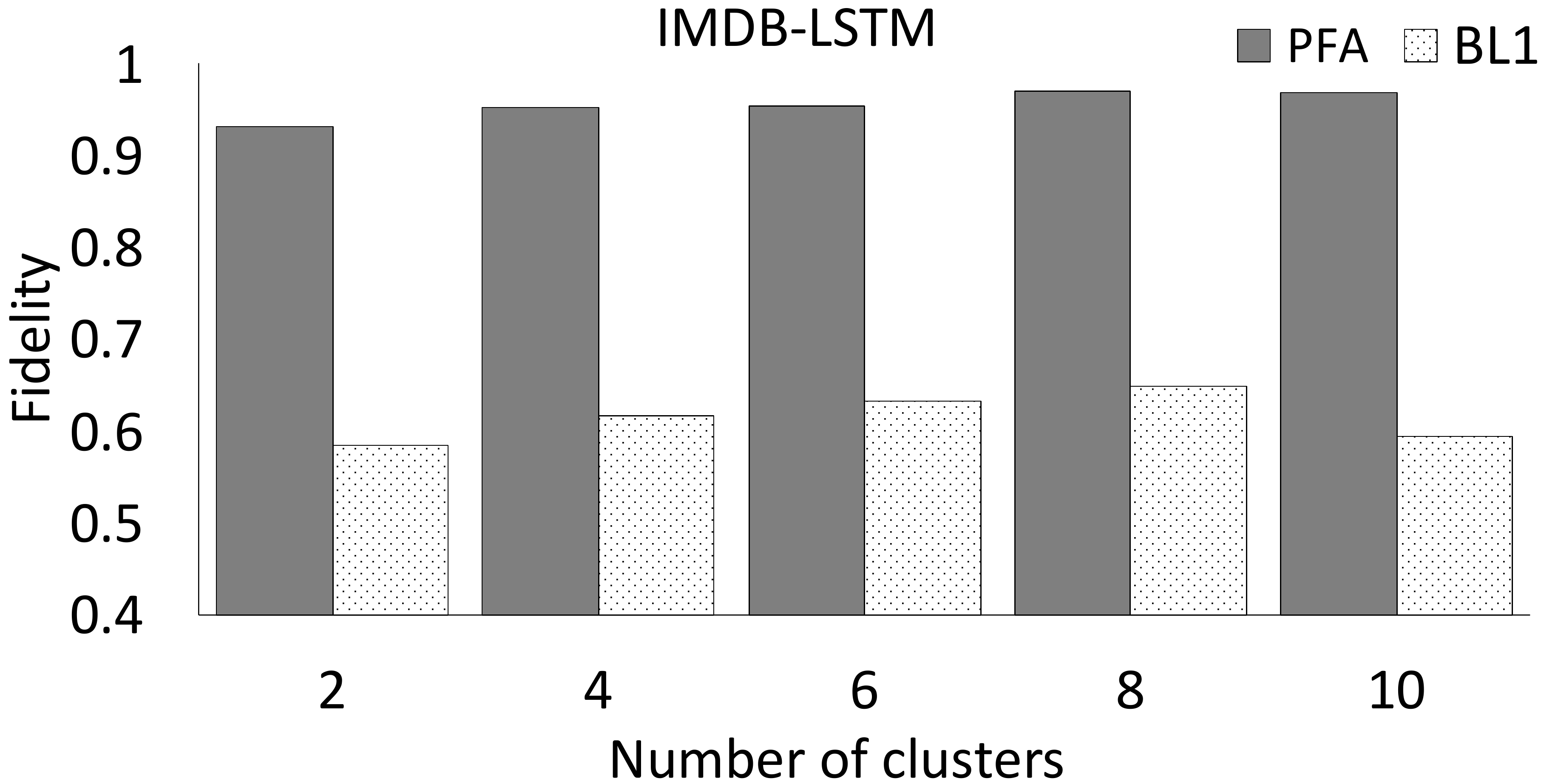} \includegraphics[width=0.24\textwidth]{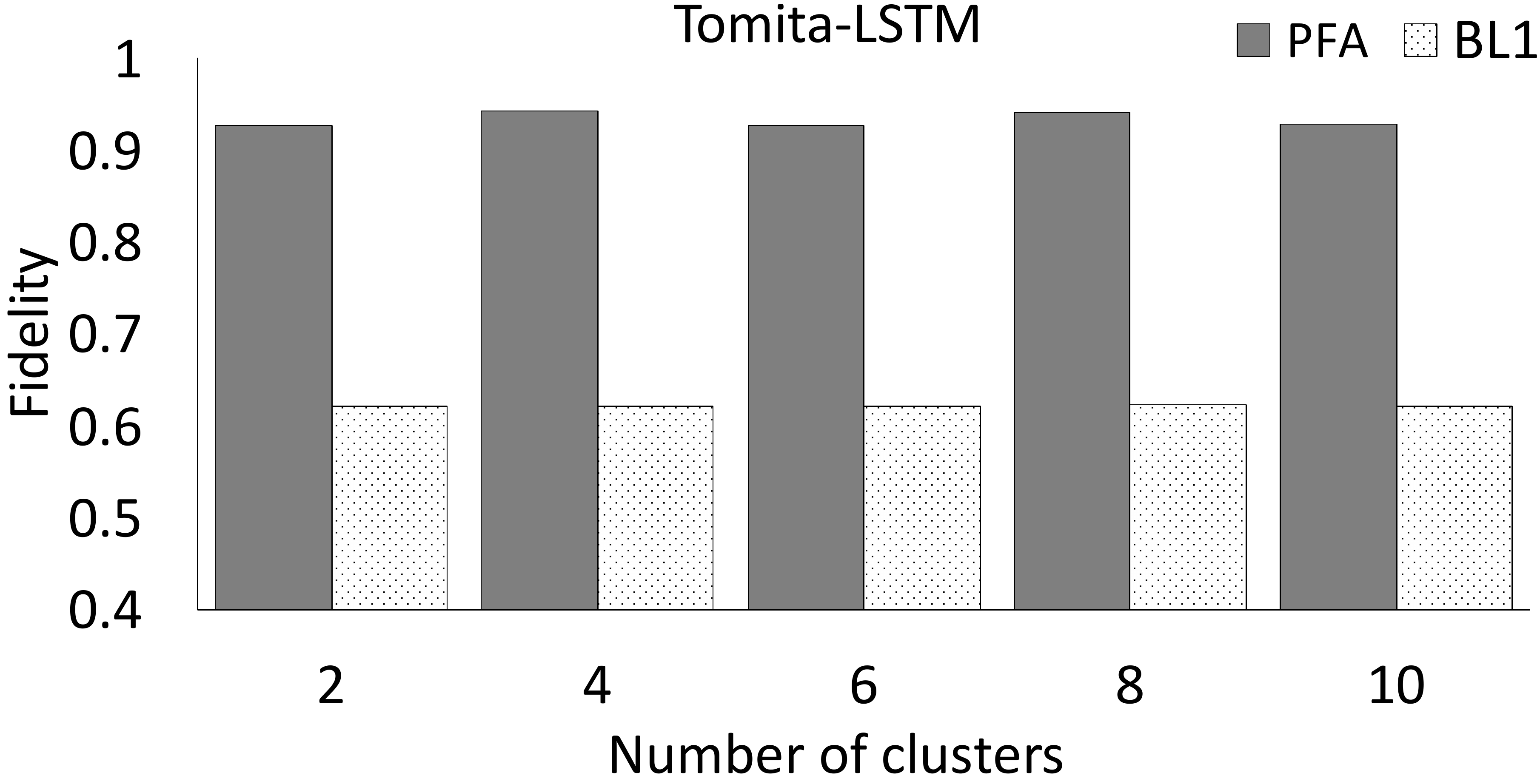} \includegraphics[width=0.24\textwidth]{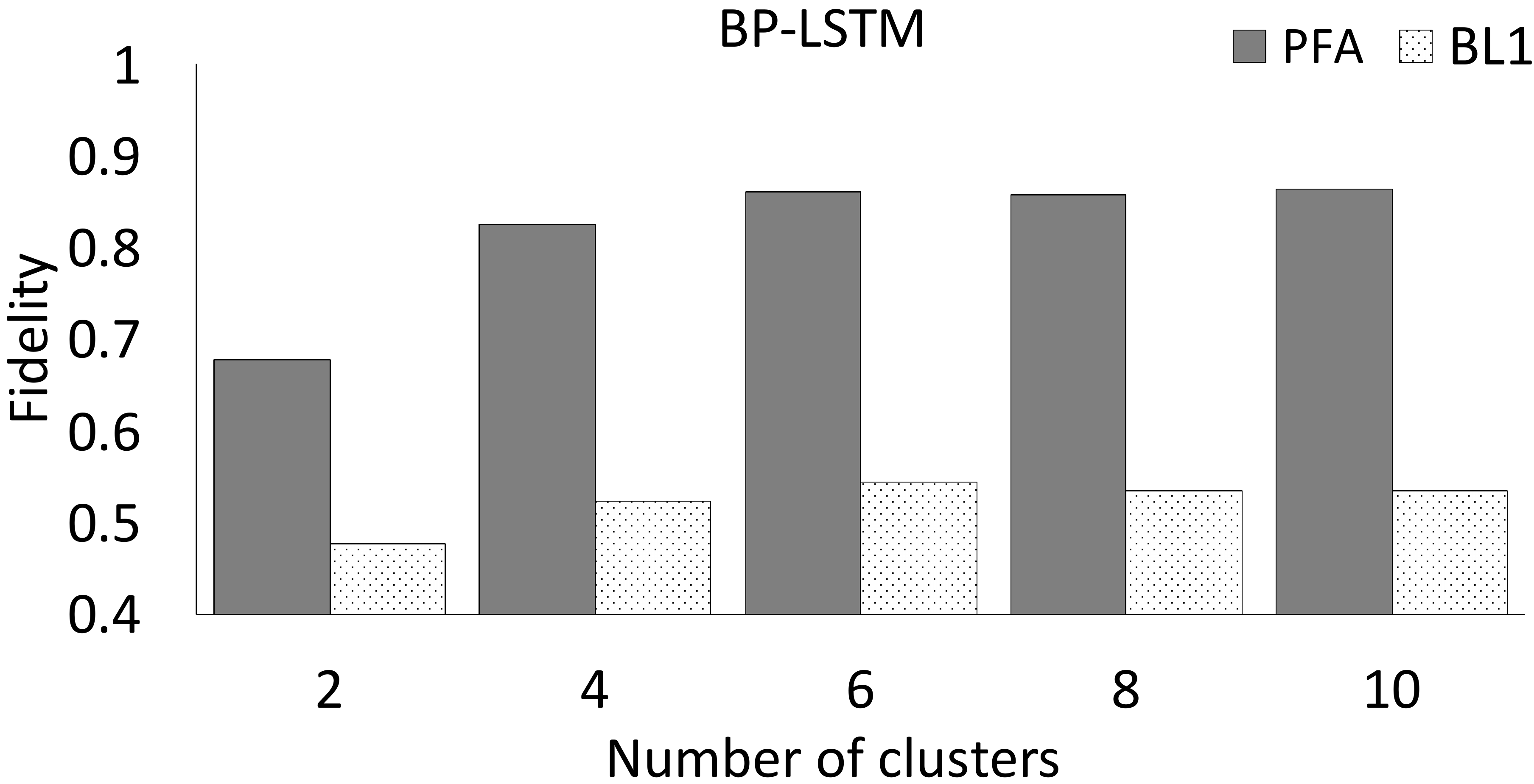} \end{subfigure}
   \begin{subfigure}{\textwidth}
             \centering
             \includegraphics[width=0.24\textwidth]{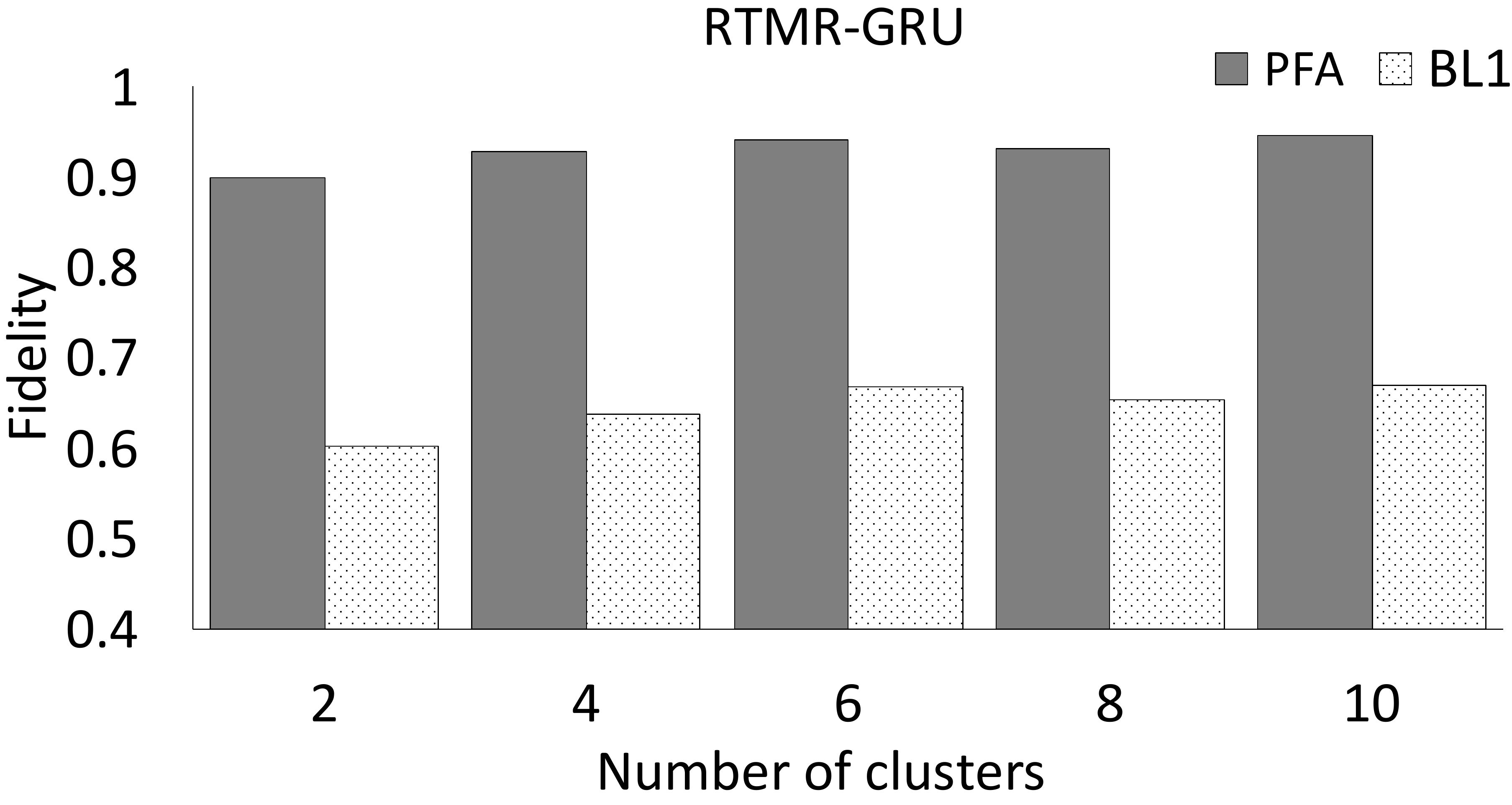} \includegraphics[width=0.24\textwidth]{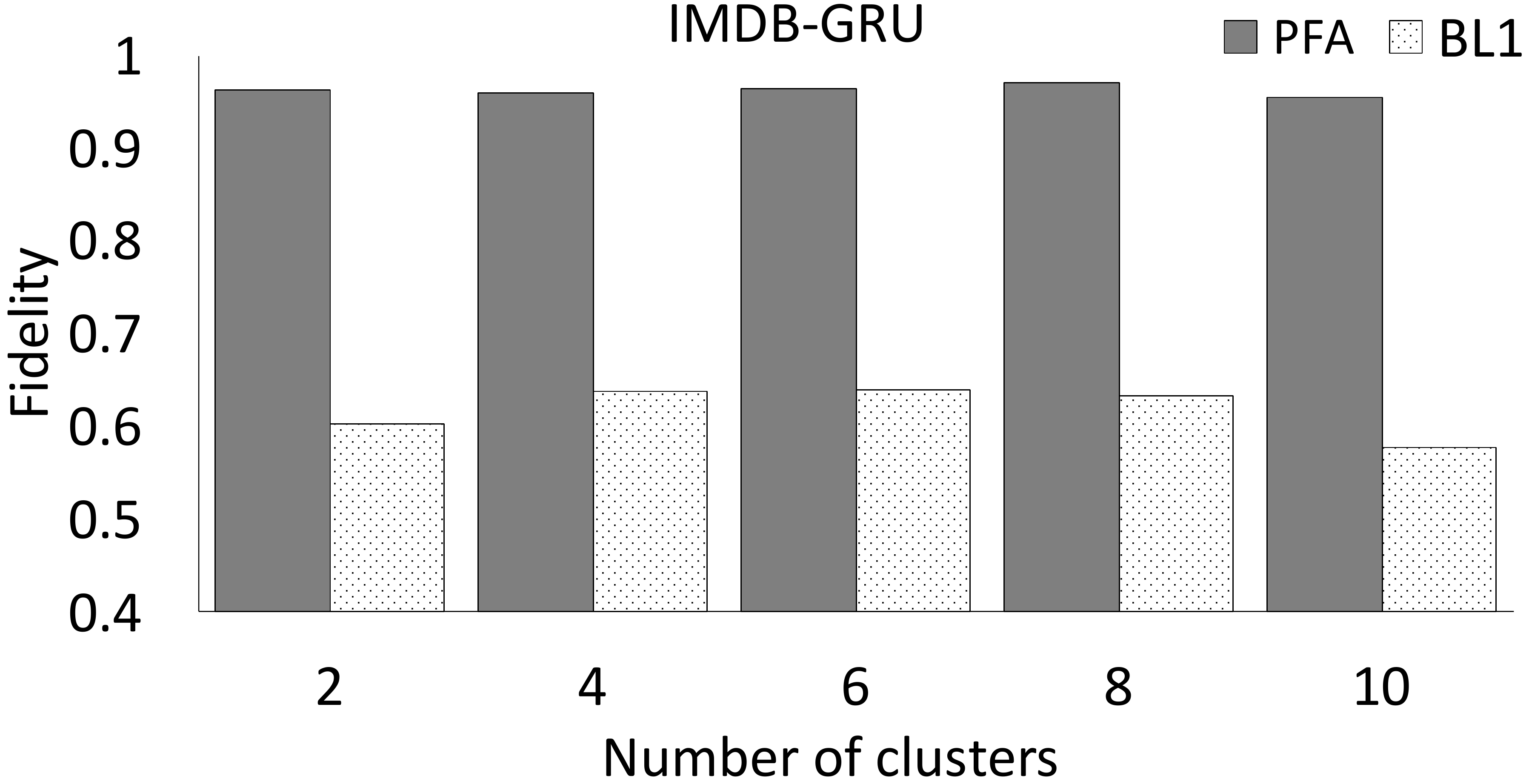} \includegraphics[width=0.24\textwidth]{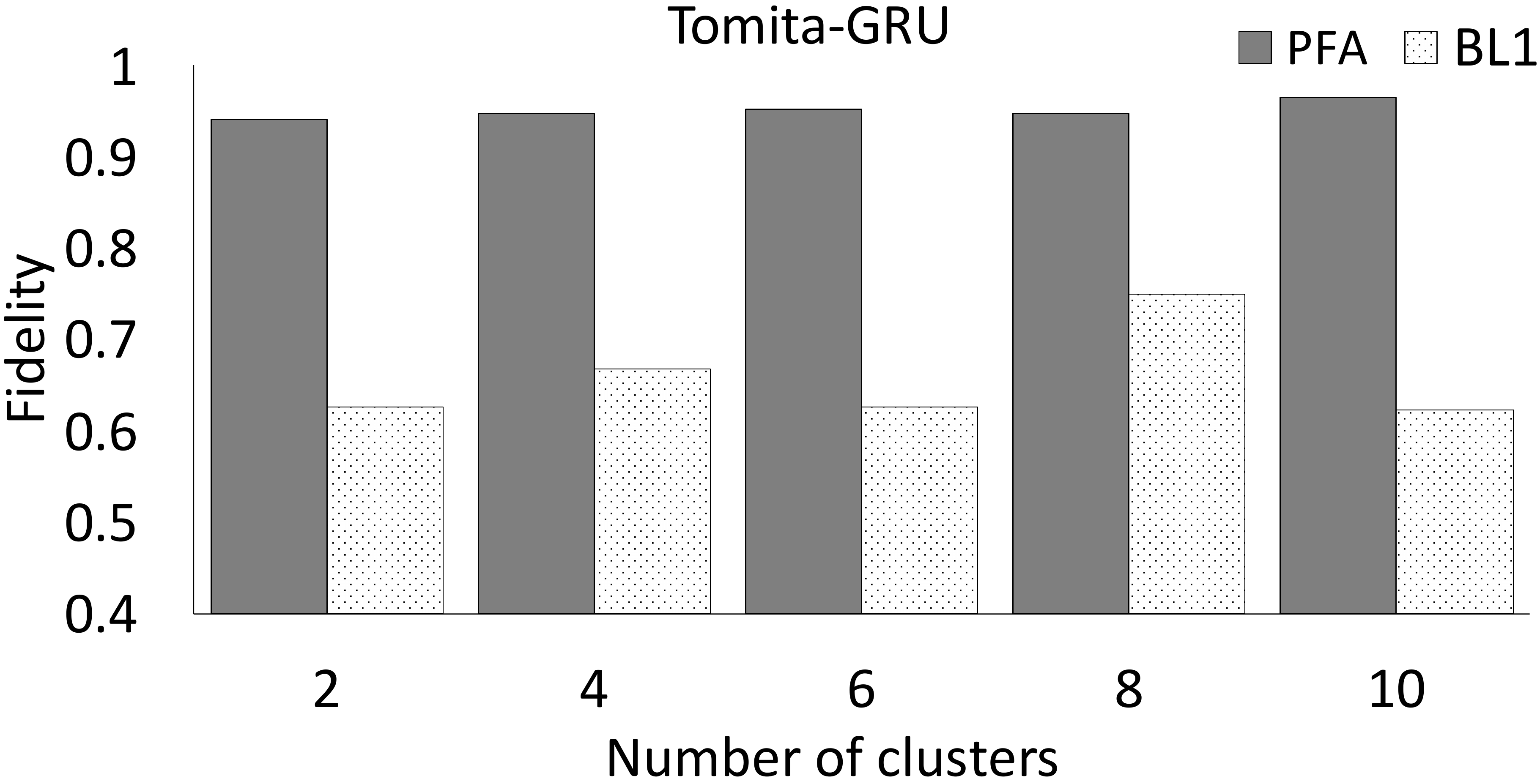} \includegraphics[width=0.24\textwidth]{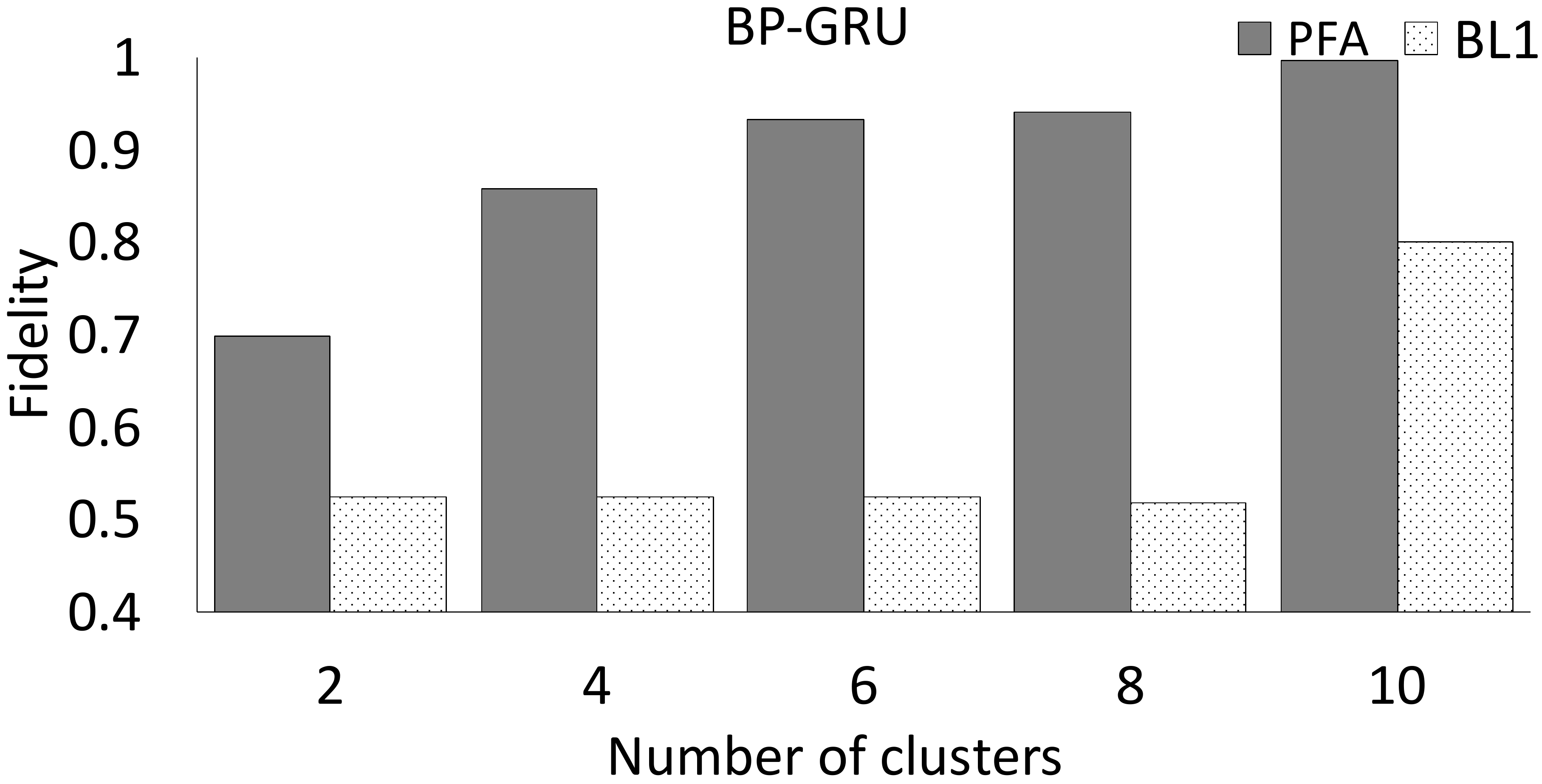} \end{subfigure}
   \caption{The fidelity of our approach vs. BL1}
    \label{fig:fid}
 \end{figure*}
 
We apply our approach to all 20 models to extract models. We evaluate whether the models precisely approximate the RNNs using two measurements, i.e., accuracy and fidelity. The former measures the percentage of the samples in the test set for which the extracted model generate the correct label. The latter measures how consistent the extracted model and the RNN model are, which is defined as follows.
\begin{equation}
\label{eq:fid}
  Fidelity=\frac{\sum_{x \in T}Sign(\mathcal{A}(x)=R(x))}{|T|}
\end{equation}
where $\mathcal{A}$ is the extracted model; $R$ is the RNN; $T$ is the test set; $x$ is any sample in the test case; and $Sign(y)$ is a sign function which equals 1 if $y$ holds and 0 otherwise. In the following, we discuss the experiment results via a comparison with existing approaches. There are three approaches which we can potentially compare to. \\

\noindent \emph{First baseline}
The first one is the approach proposed in~\cite{zhou2018learning} (referred to as BL1), which extracts DFA from RNN. They similarly reduce the hidden state space through clustering and then regard each cluster as a state of the learned automaton.
Next, they map the hidden state trace of each input into an abstract trace. Finally, the transitions between states in the abstract state traces that occur the most frequent are taken as transitions in the learned automaton. 

For a systematic comparison with BL1, we vary the number of clusters used for abstracting the hidden states for both approaches. Figure~\ref{fig:acc} shows the accuracy of extracted models using BL1 and our approach respectively. Notice that the results on the Tomita Grammars are the average of the 7 grammars for the sake of space. It can be observed that the models extracted with our approach are significantly more accurate than those generated by BL1 across all 20 models. While our approach consistently achieves an accuracy of 70\% to nearly 90\%, BL1's accuracy ranges from 50\% to slightly above 60\%.
This is expected as the models that BL1 extracts only contain transitions with maximum frequency while our approach is able to preserve all transitions through a probability distribution. Furthermore, it can be observed that in most cases the models extracted using our approach have a performance similar to that of the original RNN models, i.e., most of the extracted models have an accuracy within 10\% difference with the original models. This suggests that our approach is capable of extracting precise models which have similar performance with the RNN.

\begin{table*}[t]
  \centering
  \caption{Accuracy comparison between our approach and BL2}
    \begin{tabular}{|c|c|c|c|c|c|c|c|c|c|c|c|}
    \hline
    Model & Method & Tomita1 & Tomita2 & Tomita3 & Tomita4 & Tomita5 & Tomita6 & Tomita7 & BP & RTMR & IDMB \\
    \hline
    \multirow{2}[4]{*}{LSTM} & Ours   &0.9872 & 0.9625  &0.913 &0.9697 & 0.7487 & 0.6167&  0.9513 & 0.8642 & 0.7601 & 0.8264\\
				       & BL2   & 0.9872 & 0.9812  & 1          &1          & 0.7452 &0.6420& 1     & 0.997 & timeout & timeout \\
    \hline
    \multirow{2}[4]{*}{GRU} & Ours   &0.9872 & 0.9625 & 0.942 & 0.967 & 0.7227 &  0.6135  & 0.9875 & 0.9869 &0.7629  &0.7964  \\
				      & BL2   & 0.9872 & 0.9812  & 1          & 1          & 0.7522 & 0.6441 & 1    & 0.997 & timeout & timeout \\
    \hline
    \end{tabular}\label{tab:bl2}\vspace{-0.5em}
\end{table*}

\begin{table*}[t]
  \centering
  \caption{Fidelity comparison between our approach and BL2}
    \begin{tabular}{|c|c|c|c|c|c|c|c|c|c|c|c|}
    \hline
    Model & Method & Tomita1 & Tomita2 & Tomita3 & Tomita4 & Tomita5 & Tomita6 & Tomita7 & BP & RTMR & IDMB \\
    \hline
    \multirow{2}[4]{*}{LSTM} & Ours   & 1 &0.9563 & 0.9144 & 0.9697 & 0.9931 & 0.8648 & 0.9513  &0.8642 &0.9349&0.9528\\
				       & BL2   & 1& 0.9938 & 0.9986 & 1& 0.9757&  0.9007&  1 &0.997& timeout & timeout \\
    \hline
    \multirow{2}[4]{*}{GRU} & Ours   & 1 &0.9563&  0.9434&  0.967 &0.9584&  0.8807&  0.9875&  0.9879 &   0.9203 & 0.8553 \\
				      & BL2   & 1 &0.9938&  0.9986&  1& 0.9775&  0.905& 1& 0.996 & timeout & timeout \\
    \hline
    \end{tabular}\label{tab:bl2-fdlt}\vspace{-1em}
\end{table*}

In terms of fidelity, as shown in Figure~\ref{fig:fid}, it can be observed that our approach is significantly better than BL1 as well. The fidelity of the models extracted using our approach  ranges from 82\%(BP-LSTM) to over 95\%, whereas that of the models extracted using BL1 ranges from 52\%(BP-LSTM) to about 65\% only. Specifically, the fidelity comparison for Tomita grammars, BP, RTMR and IMDB are 94\% vs. 64\%, 85\% vs. 55\%, 89\% vs. 64\% and 96\% vs. 62\% respectively. The differences are more noticeable for real-world complex tasks like IMDB. One possible reason is that the real-world datasets are complicated and the idea underling BL1 does not apply in the real-world setting. In comparison, our probabilistic abstraction approach is capable of taking into consideration the probability distribution among the abstract states and thus extract accurate models. Note that our approach extracts models with high fidelity, i.e., most of the models have fidelity larger than 90\%, which shows that our extracted models often precisely approximate the RNNs. \\

\noindent \emph{Second baseline} The second approach we compare to is the one in~\cite{weiss2017extracting} (referred to as BL2), which applies the L* algorithm~\cite{angluin1987learning} to extract a DFA from RNN. It first builds a DFA based on an observation table, and then builds an abstract DFA from the RNN with an interval partition function (i.e., a heuristic-based abstraction). After that, it checks the equivalence between the two DFAs and refines one of them if a conflict occurs. The algorithm repeats the above procedure until the two DFAs are equivalent and returns the DFA. We remark that checking equivalence of two DFAs is expensive and impractical when the alphabet is large which is the case for real-world tasks.

Since there is no clustering in BL2,
we apply Algorithm~\ref{alg:overall} with a fidelity requirement of $\gamma_a = 0.99$ and a timeout of 400 seconds. Note that the same timeout is set for BL2, which is also the one adopted in~\cite{weiss2017extracting}.  The results in terms of accuracy are shown in Table~\ref{tab:bl2}. First, it can be observed that BL2 fails to work on either the RTMR or IMDB dataset. This is expected as the alphabet of these datasets is the vocabulary of the training set, which is 20995 for RTMR and 388441 for IMDB. They are thus way beyond the capability of BL2. Second, our approach achieves competitive results with BL2 on the two artificial datasets, i.e., on average, BL2 has an accuracy that is 3.32\% more than our approach. The results of comparing fidelity are shown in Table~\ref{tab:bl2-fdlt}. We can observe that our approach achieves high fidelity, i.e., 94.29\% on average, with the RNN model on all dataset and BL2 fails to report the results on both RTMR and IMDB for the same reason. On the two artificial datasets, BL2 has a fidelity that is 3.39\% more than our approach on average.  \\

\noindent \emph{Third baseline} The third approach is the recent approach reported in~\cite{DBLP:conf/nips/WeissGY19}. It learns a probabilistic model for approximating RNN through an extended version of the L* algorithm. While it has impressive performance on tasks with a small alphabet (like in the case of the two artificial datasets), the authors admittedly report~\cite{DBLP:conf/nips/WeissGY19} that their approach does not apply when the alphabet is large (like in the case of the two real-world datasets). This is confirmed in our experiments as well, i.e., their implementation failed to work on either RTMR or IMDB. We omit a detailed comparison due to its limited applicability and the fact that it is not implemented for classification tasks, which makes it hard to compare to.\\

\noindent \emph{Based on the above experiment results, we thus conclude that our approach is able to extract accurate models from RNN and is capable of handling real-world RNN models.
}
\vspace{-1em}
\subsection{Level of Abstraction}
Our approach allows users to specify a target fidelity and aims to extract a model based on the right level of abstraction. This is done through controlling the number of clusters, which determines the size of the alphabet and consequently the size of the extracted models. Algorithm~\ref{alg:overall} is designed based on the hypothesis that the more clusters we use, the more fine-grained the abstraction is and thus the more accurate the extracted model will be. Since the more clusters we use, the more complicated (i.e., the less comprehensive) the extracted model will be, it is important to find a balance.

To evaluate whether this hypothesis holds and understand the relationship between the number of clusters and the fidelity/size of the extracted models, we conduct the following experiments. We systematically extract models with clusters ranging from 2 to 10 and evaluate the size and accuracy/fidelity of the extracted models. Table~\ref{tab:rq3} summarizes how the size of extracted PFA changes with different numbers of clusters. Note that these results are based on applying our approach to the GRU models. Similar results are obtained on the LTSM models and are thus omitted. We observe that as we increase the number of clusters, the size of the extracted models increases in most of the cases. However, it is not monotonically so. For instance, for the BP dataset, the number of states decreases when the number of clusters increases from 6 to 10. This is because the number of states is determined jointly by the number of clusters and the degree of generalization achieved by Algorithm~\ref{alg:aa}. It is thus possible that in some cases, the same cluster (of hidden feature values) behaves differently in different contexts (e.g., the sequence of feature values before reaching the cluster) and thus must be differentiated into multiple states in the extracted PFA.

\begin{figure}[t]
\centering
\begin{minipage}{0.36\textwidth}
        \centering
        \includegraphics[width=\textwidth]{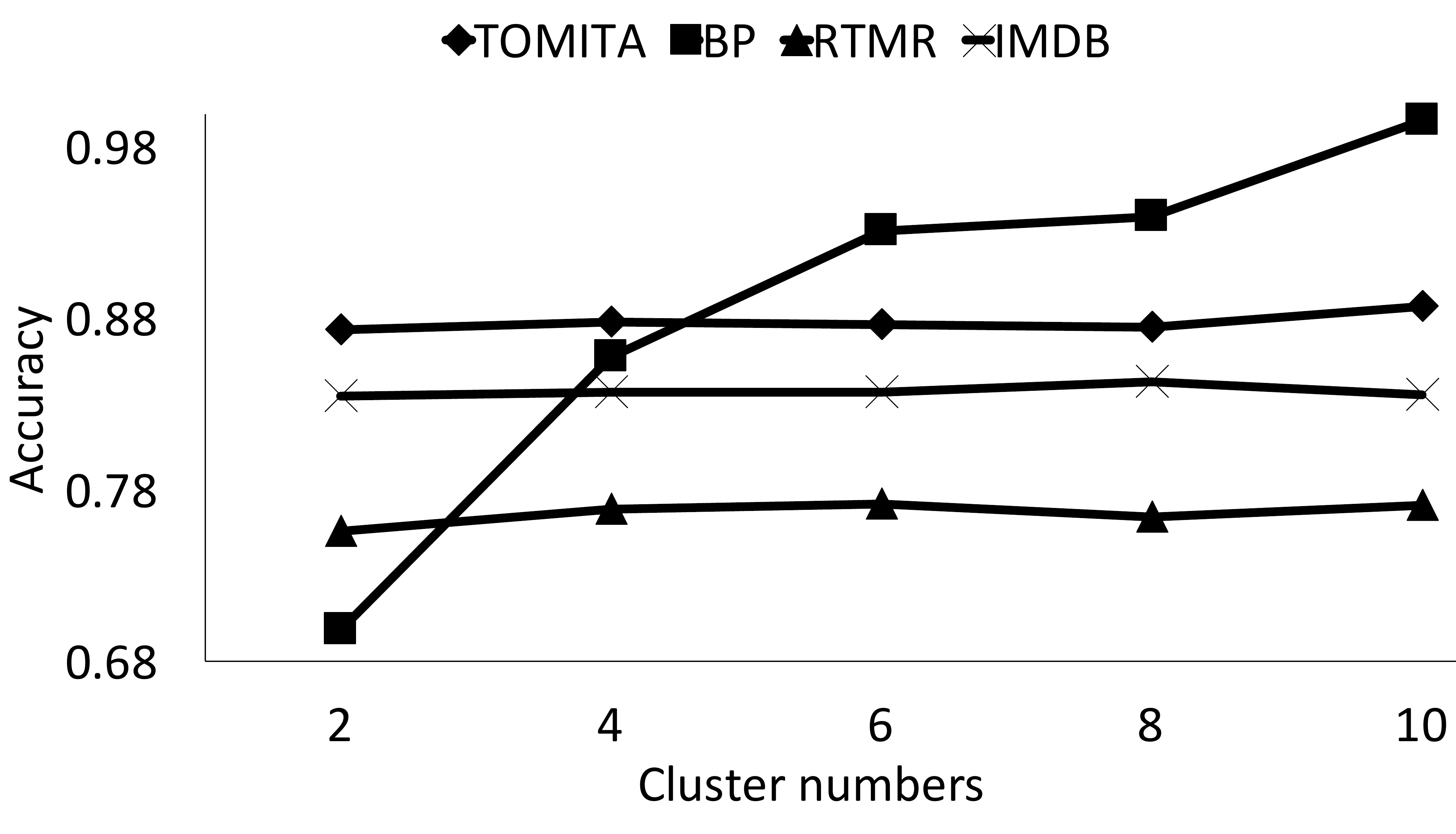} \end{minipage}
    \\
    \begin{minipage}{0.36\textwidth}
        \centering
        \includegraphics[width=\textwidth]{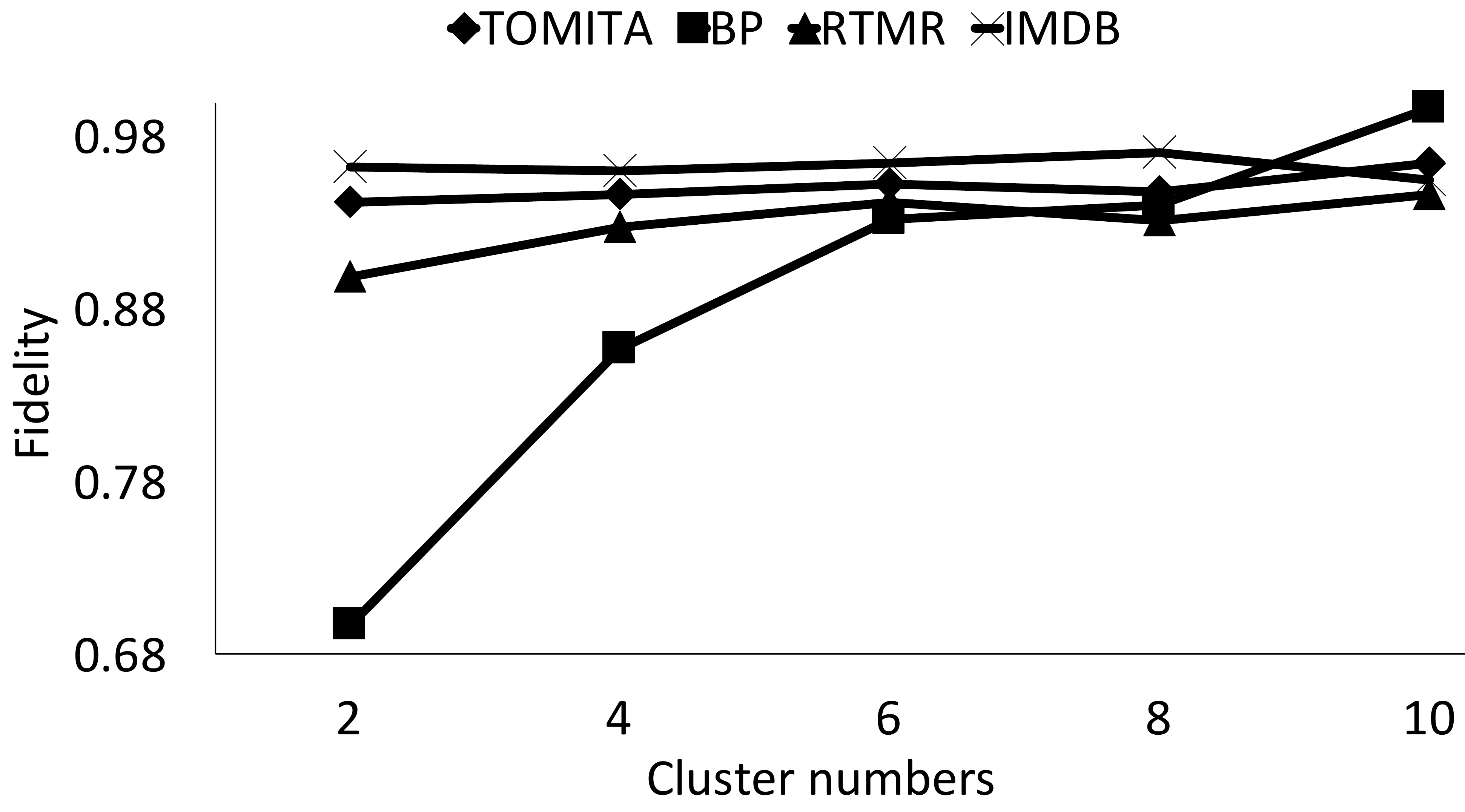} \end{minipage}
   \caption{Effects of different number of clusters.}
   \label{fig:rq3}
\end{figure}

Figure~\ref{fig:rq3} shows the relationship between the number of clusters and the accuracy/fidelity of the extracted models. We observe that as we increase the number of clusters, the accuracy/fidelity of the extracted models improves overall. The improvement, however, may vary across different models or different number of clusters. For some models, the improvement is consistent and significant, e.g., in the case of the BP dataset; for some models, the improvement is consistent but minor, e.g., in the case of Tomita grammars and RTMR; and for some models, the accuracy/fidelity may drop along the way, e.g., in the case of IMDB. For the last case, we suspect that it is due to the fact the RNN model is very complicated and Algorithm~\ref{alg:aa} failed to converge to a concise/accurate model as we notice that the number of states increases dramatically when we increase the number of clusters. This suggests a future research direction on developing new learning algorithms for probabilistic models that are effective with a large alphabet and complicated probabilistic distribution. Note that existing work such as the one in~\cite{weiss2017extracting,DBLP:conf/nips/WeissGY19} is limited to models with very small alphabets.

\begin{table}[t]
\centering
\caption{The model size with different numbers of clusters}
\begin{tabular}{|l|l|l|l|l|l|}
\hline
\multirow{2}{*}{Dataset}  & \multicolumn{5}{c|}{Number of cluster}                                              \\ \cline{2-6}
                                                                & 2  & 4  & 6   & 8   & 10 \\ \hline
Tomita & 5 & 7 & 13  & 12  & 16   \\ \hline
BP                                                              				   & 5 & 10 & 26  & 15  & 13   \\ \hline
RTMR                                                             				   & 5  & 11  & 18 & 12 & 48 \\ \hline
IMDB                                                          	 			   & 5  & 46 & 46& 567 & 1505 \\ \hline				
\end{tabular}
\label{tab:rq3}
\vspace{-1.5em}
\end{table}

\subsection{Usefulness}
We have shown that our approach is able to extract models which approximate RNN accurately. This facilitates some degree of human interpretation and automatic analysis of RNN. For instance, given an RNN trained for sentiment analysis, with the extracted model, we can systematically compute the probability of generating a `positive' label (through manual computation if the model is very small or probabilistic model checking otherwise) after provided with each word in a sentence. By monitoring how the probability changes with each word, we can develop some intuitive understanding on how the sentiment analysis result is derived. Such usefulness is, however, subjective. In the following, we report an application of the extracted PFA models for adversarial text detection.

Given an RNN model, adversarial texts are texts which are crafted specifically to induce mistake (i.e., so that the RNN's classification result is wrong). It has been shown in~\cite{hotflip, textbugger} that adversarial texts can be systematically generated by applying a small perturbation to a benign text (which otherwise is correctly classified by the RNN). Typical ways of generating adversarial texts include identifying and replacing important words
in a sentence with its synonyms~\cite{textbugger} or applying Neural Machine Translation twice (e.g., from English to French and then back) to the given sentence. Detecting adversarial texts is highly nontrivial and to the best of our knowledge, there have not been systematic methods proposed for that.

In the following, we show that the PFA models extracted using our method can be used to detect adversarial texts effectively. The intuition is that, given a benign text, our PFA would associate a much higher probability with its label (predicted by the RNN) than other labels; and given an adversarial text, the probability associated with each label would not be very different. This intuition is partly based on the fact that these adversarial texts are typically generated by perturbing a benign text just enough to across the classification boundary.
Based on this intuition, we design the following metric to detect adversarial texts. Given a text $x$ (which could be benign or adversarial), let
\begin{equation}
\label{eq:adv}
T(x)=\frac{P(x,y)}{P(x, \overline{y})}
\end{equation}
where $y$ is the label predicted by the RNN, $P(x,y)$ is the probability of reaching label $y$ based on our extracted PFA (which is obtained as explained in Section~\ref{sec:modelselection} using probabilistic model checking) and $P(x, \overline{y})$ denotes the summed probability of reaching labels other than $y$.
We then distinguish adversarial texts from benign ones using a threshold on $T(x)$, i.e., a text is considered as adversarial if it has a $T(x)$ smaller than the threshold.

We evaluate the effectiveness of the above method for adversarial text detection on the two real world datesets. Concretely, for each benign text in the test set of RTMR and IMDB datasets, we generate an adversarial text using TEXTBUGGER~\cite{textbugger}. We then randomly select 1000 benign texts and 1000 corresponding adversarial texts to compose a test set for our detection method.
We calculate $T(x, y)$ for all the texts in the test set and report the AUC (Area Under Curve) score~\cite{fawcett2006introduction} to measure the effectiveness of our detection method since AUC averts the supposed subjectivity when selecting the threshold for a classifier and measures how true positive rate and false positive rate trade off. To further study the effect of having a different number of clusters, we apply the method with PFA extracted with different number of clusters.
The results are summarized in Table~\ref{tab:rq4} where we vary the number of clusters from 2 to 10. We observe that our method effectively detects adversarial texts, i.e., achieving an average AUC of 0.85 and 0.93 for RTMR and IMDB respectively. We do also notice that the AUC varies with the number of clusters in a way with no clear correspondence with the PFA's accuracy/fidelity, which we will investigate in the future.

The above study suggests that our model extraction approach not only offers a way of shedding some light on how RNN works but also potentially opens the door for applying software analysis techniques (like model-based testing, model checking, runtime monitoring and verification) to real-world RNN models.

\begin{table}[t]
  \centering
  \caption{AUC of adversarial sample detection.}
    \begin{tabular}{|c|c|c|c|c|c|c|}
      \hline
    \multirow{2}[4]{*}{Dataset} & \multirow{2}[4]{*}{Model} & \multicolumn{5}{c|}{Number of clusters} \\ \cline{3-7}
      &       & 2     & 4     & 6     & 8     & 10 \\     \hline
    \multirow{2}[4]{*}{RTMR} & LSTM  & 0.5745 & 0.8101 & 0.8328 & 0.8453 & \textbf{0.8479} \\ \cline{2-7}
     				        & GRU   & 0.6794 & 0.7800 & 0.8183 & 0.8409 & \textbf{0.8459} \\     \hline
    \multirow{2}[4]{*}{IMDB} & LSTM  & 0.7670 & 0.8771 & 0.8949 & \textbf{0.9274} & 0.9173 \\ \cline{2-7}
	& GRU   & 0.7014 & 0.8756 & 0.9163 & \textbf{0.9323} & 0.9228 \\
        \hline
    \end{tabular}\label{tab:rq4}\end{table}\vspace{-0.64cm}

   \vspace{1.5em}
\section{Related Work}
\label{sec:re}
We review related works in this section.
From a broader point of view, this work is relevant to the explanation of machine learning models which can be categorized into local and global explanations in general. Intuitively, local explanation tries to explain why the target machine learning model makes a decision on a certain input. One example is the SHAP-like system~\cite{shap}, which uses a linear function to mimic the complex models, e.g., convolutional neural network (CNN), when producing a certain output on an input. Global explanation, however, aims to understand the internal decision process by using a more interpretable model to mimic the behaviors of the original model on \emph{any} inputs. One example is the work in~\cite{jacobsson2005rule}. Our work takes a global explanation perspective.

This work is related to work on RNN rule extraction. Rule extraction from RNN is the process of constructing different computational models which mimic the RNN~\cite{jacobsson2005rule,ayache2018explaining}. This work is especially related to the work that extracts a deterministic finite automaton (DFA) from RNN. These approaches usually rely on encoding the hidden states into symbolic representations using techniques like clustering~\cite{zhou2018learning} or interval partitioning~\cite{weiss2017extracting}. Our work is different by learning a probabilistic finite automaton (PFA) from the symbolic data. There is also some recent work aiming to extract a weighted automaton (WA)~\cite{ayache2018explaining} or discrete-time Markov Chain~\cite{Du:2019:DMQ:3338906.3338954}. However, neither of them provide generalization to capture the temporal dependency over the symbolic representations. Our work encodes the concrete states in a similar way but then uses probabilistic abstraction to extract a probabilistic model.

The study of learning PFA is a branch of grammar inference~\cite{gi}, which has been investigated under different settings using methods like state merging~\cite{alergia,wang2017should} or identifying the longest dependent memory~\cite{ron1996power,ron1998learnability}. Recently, researchers have proposed to learn PFA for system analysis tasks like model checking or runtime monitoring~\cite{mao}. This work follows a state-merging learning paradigm to learn a PFA from the symbolic data extracted from RNN.
 \section{Conclusion}
\label{sec:con}
In this work, we propose to extract probabilistic finite automata from state-of-the-art recurrent neural network to trace/mimic its behaviors for analysis by probabilistic abstraction. Our approach is based on symbolic encoding of RNN hidden state vectors and a probabilistic learning algorithm which tries to recover the probability distribution of the symbolic data. The experiment results on real-world sentiment analysis tasks show that our approach significantly improves the quality or scalability of state-of-the-art model extraction works. Our approach provides one promising way to bridge the gap for applying a variety of software/system analysis techniques to real-world neural networks.

 \section*{acknowledgements}
This research is supported by the National Key R\&D Program of China under Grant No. 2019YFB1600700  and Project of Science and Technology Research and Development Program of China RailwayCorporation (P2018X002) and NSFC Program (Grant No. 61972339). This research is also supported by the National Research Foundation, Singapore under its AI Singapore Programme (AISG Award No: AISG-RP-2019-012). Any opinions, findings and conclusions or recommendations expressed in this material are those of the author(s) and do not reflect the views of National Research Foundation, Singapore. This research has also been supported by the Key-Area Research and Development Program of Guangdong Province (Grant no. 2018B010107004), the Fundamental Research Funds for the Zhejiang University NGICS Platform, and the National Research Foundation, Prime Minister’s Office, Singapore under its Corporate Laboratory@University Scheme, National University of Singapore, and Singapore Telecommunications Ltd.
\bibliographystyle{plain}
\bibliography{ref}
\end{document}